\documentclass[iicol,pdflatex]{sn-jnl}

\usepackage{graphicx}%
\usepackage{multirow}%
\usepackage{amsmath,amssymb,amsfonts}%
\usepackage{amsthm}%
\usepackage{mathrsfs}%
\usepackage[title]{appendix}%
\usepackage{xcolor}%
\usepackage{textcomp}%
\usepackage{manyfoot}%
\usepackage{booktabs}%
\usepackage{listings}%
\usepackage[square,numbers]{natbib}
\usepackage{color, soul} % highlight: \hl{This will be highlight.}
\sethlcolor{yellow}
\soulregister\cite7 
\soulregister\citep7 
\soulregister\citet7 
\soulregister\ref7 
\soulregister\pageref7 

\usepackage{tabularx}
\usepackage[ruled, vlined, linesnumbered]{algorithm2e}
\usepackage{siunitx}

\usepackage{tablefootnote}
\usepackage{threeparttable}
\theoremstyle{thmstyleone}%
\theoremstyle{thmstyletwo}%

\theoremstyle{thmstylethree}%

\begin{document}

\title[Reinforced Imitative Trajectory Planning for Urban Automated Driving]{Reinforced Imitative Trajectory Planning for Urban Automated Driving}


\author[1]{\fnm{Di} \sur{Zeng}}\email{zigned@cqu.edu.cn}

\author*[1,2]{\fnm{Ling} \sur{Zheng}}\email{zling@cqu.edu.cn}

\author[1]{\fnm{Xiantong} \sur{Yang}}\email{xiantongyang@cqu.edu.cn}

\author[1,2]{\fnm{Yinong} \sur{Li}}\email{ynli@cqu.edu.cn}

\affil[1]{\orgdiv{College of Mechanical and Vehicle Engineering}, \orgname{Chongqing University}, \orgaddress{\street{Shazheng Street}, \city{Chongqing}, \postcode{40044}, \state{Chongqing}, \country{China}}}
\affil*[2]{\orgdiv{State Key Laboratory of Mechanical Transmission for Advanced Equipment}, \orgname{Chongqing University}, \orgaddress{\street{Shazheng Street}, \city{Chongqing}, \postcode{40044}, \state{Chongqing}, \country{China}}}

%\author*[1,2]{\fnm{First} \sur{Author}}\email{iauthor@gmail.com}
%\affil*[1]{\orgdiv{Department}, \orgname{Organization}, \orgaddress{\street{Street}, \city{City}, \postcode{100190}, \state{State}, \country{Country}}}
%
%\affil[2]{\orgdiv{Department}, \orgname{Organization}, \orgaddress{\street{Street}, \city{City}, \postcode{10587}, \state{State}, \country{Country}}}
%
%\affil[3]{\orgdiv{Department}, \orgname{Organization}, \orgaddress{\street{Street}, \city{City}, \postcode{610101}, \state{State}, \country{Country}}}


\abstract{Reinforcement learning (RL) faces challenges in trajectory planning for urban automated driving due to the poor convergence of RL and the difficulty in designing reward functions. Consequently, few RL-based trajectory planning methods can achieve performance comparable to that of imitation learning-based methods. The convergence problem is alleviated by combining RL with supervised learning. However, most existing approaches only reason one step ahead and lack the capability to plan for multiple future steps. Besides, although inverse reinforcement learning holds promise for solving the reward function design issue, existing methods for automated driving impose a linear structure assumption on reward functions, making them difficult to apply to urban automated driving. In light of these challenges, this paper proposes a novel RL-based trajectory planning method that integrates RL with imitation learning to enable multi-step planning. Furthermore, a transformer-based Bayesian reward function is developed, providing effective reward signals for RL in urban scenarios. Moreover, a hybrid-driven trajectory planning framework is proposed to enhance safety and interpretability. The proposed methods were validated on the large-scale real-world urban automated driving nuPlan dataset. Evaluated using closed-loop metrics, the results demonstrated that the proposed method significantly outperformed the baseline employing the identical policy model structure and achieved competitive performance compared to the state-of-the-art method. The code is available at \url{https://github.com/Zigned/nuplan_zigned}.}




\maketitle

\section{Introduction}\label{sec_introduction}

Urban automated driving (AD) presents significant challenges due to the complexity and diversity of traffic scenarios. To ease the difficulty, automated driving systems are usually divided into modules, such as perception, prediction, planning, and control. The planning module plays a crucial role in achieving a safe and comfortable driving experience since it determines the behavior of automated vehicles (AVs). Despite the interpretability, conventional approaches for planning heavily rely on hand-crafted rules, making it difficult to handle complex and diverse situations. In contrast, learning-based approaches learn planning policies from large-scale data, thus circumventing the need for meticulously designed rules.

In recent years, imitation learning (IL) and reinforcement learning (RL) have dominated the field of planning. IL learns the optimal planning policy from expert demonstrations. The most common category of IL is behavior cloning (BC), where expert demonstrations serve as supervisory signals to the agent. For example, the famous end-to-end framework proposed by NVIDIA maps raw camera inputs directly to steering commands \cite{bojarski2016end}. In Ref. \cite{Chen_2022_CVPR}, sensor inputs of different modalities were utilized to enhance the urban driving ability of the BC-based planning model. More recently, the unified AD framework, named UniAD, incorporates perception and prediction tasks in one network to improve planning performance \cite{Hu_2023_CVPR}. However, BC simply imitates the expert demonstrations in the collected states, resulting in poor generalization ability of the learned policy in new states, known as covariate shift \cite{pmlr-v9-ross10a}. To address this issue, DAgger employs the expert policy to take as input the visited state and output actions, which are aggregated into the training dataset \cite{pmlr-v15-ross11a}. However, DAgger and its extensions, such as SafeDAgger \cite{Zhang_Cho_2017}, require access to the expert policy during training, which is non-trivial.

RL is another promising solution for planning policy learning. In RL, the agent continuously interacts with the environment and optimizes its policy according to the rewards received from the environment. Unlike BC, RL trains the agent in a closed-loop manner, enabling the agent to continuously accumulate driving experience and understand the closed-loop effects of its actions \cite{Zhang_Guo_2022}. 

RL has been successfully applied in highway driving tasks, such as lane keeping \cite{sallab2016end,8793742}, lane changing \cite{9978654, 9325948}, and traffic merging \cite{8317735}, where the agents are usually responsible for steering, throttle, and braking control according to vectorized traffic information (e.g., relative positions and relative velocities). While RL has also been explored for urban scenarios, these environments involve more complex situations than highway scenarios, attributed to intricate road structures and diverse traffic participants with different behavioral patterns. This complexity necessitates more sophisticated state representations, leading to larger parameter spaces and more complex neural network architectures for RL agents. Consequently, the gradients computed by RL become insufficient to optimize such deep neural networks \cite{chen2023end}. To address this, some approaches have combined RL with supervised learning. For instance, an IL agent was pre-trained to regularize the RL agent training by imposing the Kullback-Leibler divergence between the two agents' policies in Ref. \cite{9694460}. The RL agent determined a longitudinal target speed and a target lane according to bird's-eye-view (BEV) rasterized images, which incorporate map information and relative states between traffic participants. Roach \cite{Zhang_2021_ICCV} trains an RL agent with privileged BEV semantic segmentation images. Then, the RL agent is employed to collect a dataset, with which a downstream IL agent learns to control the steering, throttle, and braking. More recently, GIGAFLOW \cite{cusumano2025robust} enables effective training of policies via self-play reinforcement learning on a massive scale. In GIGAFLOW, all traffic participants share the same policy and collecting massive experience in parallel, which mitigates the problem of insufficient gradient information.

However, few RL-based methods for AD are competitive to IL-based methods, especially in urban scenarios \cite{chen2023end}. Beyond the aforementioned convergence challenges, another significant factor is that most existing RL methods for AD reason only one step ahead and lack the capability of planning for multiple future steps \cite{Zhang_Guo_2022, chen2023end, 9660769, 9351818}, which can be necessary to handle complex urban scenarios. To enable multi-step reasoning, TrajHF \cite{li2025finetuning} trains a diffusion model via reinforcement learning with human feedback to generate multiple future states. However, the iterative denoising process required for state generation can be computationally expensive during inference. As an alternative, trajectory value learning (TRAVL) \cite{Zhang_Guo_2022} defines the action in RL as trajectories, which circumvents the need for explicit multi-step planning by directly learning the value of entire trajectories in RL. Nevertheless, TRAVL has several limitations. First, TRAVL selects a trajectory from a finite trajectory set generated by predefined rules, which may reduce trajectory optimality. Second, generating the trajectory set and evaluating every trajectory online increase computational complexity. Third, the applied scenarios of TRAVL were highway scenarios, while urban scenarios are much more complex.

Besides, most approaches rely on hand-crafted reward functions. Although inverse reinforcement learning (IRL) holds promise for solving the reward function design issue, existing methods \cite{sharifzadeh2016learning, 9126156, 9460807} for AD mostly assume reward functions to be linear combinations of manually designed features, making them challenging to apply to urban traffic scenarios. Thus, designing effective reward functions for urban AD remains challenging.

Addressing the aforementioned limitations, this paper proposes a novel RL-based trajectory planning method, Reinforced Imitative Trajectory Planning (RITP), for urban automated driving. The contributions and novelty of this study are summarized as follows:
\begin{itemize}
	\item[1)]
	A novel RL-based trajectory planning method, RITP, is proposed. While most existing RL-based methods can only reason one step ahead, RITP integrates RL with IL, enabling planning for multiple future steps in complex urban scenarios. Unlike TRAVL, which plans trajectories with reduced optimality, RITP trains a transformer-based planner to explore the state and trajectory spaces using trajectory noise and a transformer-based value function to evaluate the closed-loop effects of a plan. After training, the planner directly outputs a trajectory given a state, thus avoiding online trajectory generation and evaluation encountered by TRAVL. 
	\item[2)]
	A transformer-based Bayesian reward function is developed. The reward function is a Bayesian neural network without the linear structure assumption and models the uncertainty of the neural network, providing effective reward signals for reinforcement learning in urban scenarios. 
	\item[3)]
	A hybrid-driven trajectory planning framework is proposed to combine the advantages of data-driven and model-driven methods. The framework utilizes the trajectory planned by the RL agent to guide numerical optimization-based trajectory refinement, assessing the safety, legality, and comfort of the planned trajectories in an interpretable way.
	\item[4)]
	The proposed methods were validated on the large-scale real-world urban automated driving nuPlan dataset \cite{caesar2021nuplan, karnchanachari2024towards}. The effectiveness of the proposed methods was evaluated in both closed-loop non-reactive and closed-loop reactive experiments.
\end{itemize}

\section{Reinforced Imitative Trajectory Planning}\label{sec_RITP}

\subsection{Preliminaries}\label{subsec_preliminaries}
AD can be modeled as a Markov decision process (MDP) defined by the tuple ($\mathcal{S}, \mathcal{A}, \mathcal{T}, R, \gamma$), with states $s \in \mathcal{S}$, actions $a \in \mathcal{A}$, a transition function $\mathcal{T}(s_{t+1} | s_t,a_t): \mathcal{S} \times \mathcal{S} \times \mathcal{A} \rightarrow [0,1]$, a reward function $R(s): \mathcal{S} \rightarrow \mathbb{R}$, and a discount factor $\gamma \in \left(0, 1\right]$. At each time step $t$, the ego agent (i.e., the AV) takes an action $a_t$ given a state $s_t$ according to its policy $\pi (s_t)$, receives a scalar reward $r_{t+1}$, and visits a new state $s_{t+1}$, resulting in an MDP sequence $\left(s_0, a_0, r_1, s_1, a_1, \ldots, r_{T_\text{m}}, s_{T_\text{m}}, a_{T_\text{m}}\right)$, where $T_\text{m}$ is the time horizon. The return is defined as the discounted cumulative reward 
\begin{align}
	G_t&=\sum_{k=t+1}^{T_\text{m}} \gamma^{k-t-1} r_{k} \notag \\
	&=\sum_{k=t+1}^{T_\text{m}} \gamma^{k-t-1} R(s_{k-1}, a_{k-1}).
\end{align}

In RL, the goal is to identify the optimal policy $\pi_\phi^*$, parameterized by $\phi$, that maximizes the expected return $\mathbb{E}_{s_k \sim p_{\pi_\phi}, a_k \sim \pi_\phi(s_k)}\left[G_{0}\right]$, where $p_{\pi_\phi}$ is the distribution of $s_k$ under policy $\pi_\phi$. In actor-critic methods, the optimal policy (i.e., the actor) can be obtained by maximizing an approximated value function (i.e., the critic). The value function is defined as $Q^\pi(s,a)=\mathbb{E}_{s_k \sim p_\pi, a_k \sim \pi(s_k)} [G_t \mid s_t=s,a_t=a]$, which can be learned with temporal difference learning based on the Bellman equation \cite{sutton1988learning}
\begin{equation}
	Q^\pi(s,a)=R(s, a) + \gamma \mathbb{E}_{s^\prime, a^\prime} \left[Q^\pi (s^\prime, a^\prime) \right]
\end{equation}
where $(s^\prime, a^\prime)$ is the subsequent state-action pair of $(s, a)$. 

For high-dimensional problems, such as AD, the value function is usually approximated by a neural network $Q_\theta(s,a)$ parameterized by $\theta$. The deep Q-learning algorithm \cite{mnih2015human} optimizes the approximator $Q_\theta(s,a)$ to match a target $q$ outputted by a secondary frozen target network $Q_{\theta^\prime} (s,a)$:
\begin{equation}
	q = R(s, a) + \gamma Q_{\theta^\prime} (s^\prime, a^\prime),
\label{critic_target}
\end{equation}
where $a^\prime \sim \pi_{\phi^\prime} (s^\prime)$, $ \pi_{\phi^\prime}$ is a secondary frozen target actor network with parameter $\phi^\prime$.  The parameters $\theta^\prime$ and $\phi^\prime$ are either periodically substituted by $\theta$ and $\phi$ or updated by a small proportion $\xi$ at each time step, i.e., $\theta^\prime \leftarrow \xi\theta + (1-\xi) \theta^\prime$ and $\phi^\prime \leftarrow \xi\phi + (1-\xi) \phi^\prime$. However, the value estimated by $Q_\theta(s,a)$ is an overestimation \cite{pmlr-v80-fujimoto18a}. Addressing this problem, the twin delayed deep deterministic policy gradient algorithm (TD3) \cite{pmlr-v80-fujimoto18a} uses two critic networks ($Q_{\theta_1}$, $Q_{\theta_2}$) with two target critic networks ($Q_{\theta^\prime_1}$, $Q_{\theta^\prime_2}$) and replaces the target in Eq. (\ref{critic_target}) by
\begin{equation}
	q = R(s,a) + \gamma \min _{i=1,2} Q_{\theta_{i}^{\prime}}\left(s^{\prime}, \pi_{\phi}\left(s^{\prime}\right)\right).
\label{clipped_double_q_learning}
\end{equation}
To further reduce the estimation variance, TD3 adds a clipped normally distributed random noise $\epsilon$ to the target policy, resulting in a variant of Eq. (\ref{clipped_double_q_learning})
\begin{equation}
	\begin{aligned}
	q & =r+\gamma \min _{i=1,2} Q_{\theta_{i}^{\prime}}\left(s^{\prime}, \pi_{\phi^{\prime}}\left(s^{\prime}\right)+\epsilon\right), \\
	\epsilon & \sim \operatorname{clip}(\mathcal{N}(0, \sigma^2),-c, c),
	\end{aligned}
	\label{q_obj}
\end{equation}
where $\sigma$ represents the standard deviation of the normal distribution and $[-c,c]$ defines the range of $\epsilon$. Given the approximated value function and mini-batch size $N$, the policy is updated according to the deterministic policy gradient
\begin{equation}
	\left.N^{-1} \sum \nabla_{a} Q_{\theta_{1}}(s, a)\right|_{a=\pi_{\phi}(s)} \nabla_{\phi} \pi_{\phi}(s)
\end{equation}
every $d$ iterations, known as the delayed policy updates.

\subsection{Framework}\label{subsec_framework}
\begin{figure*}[h]
\centering
\includegraphics[width=0.7\textwidth]{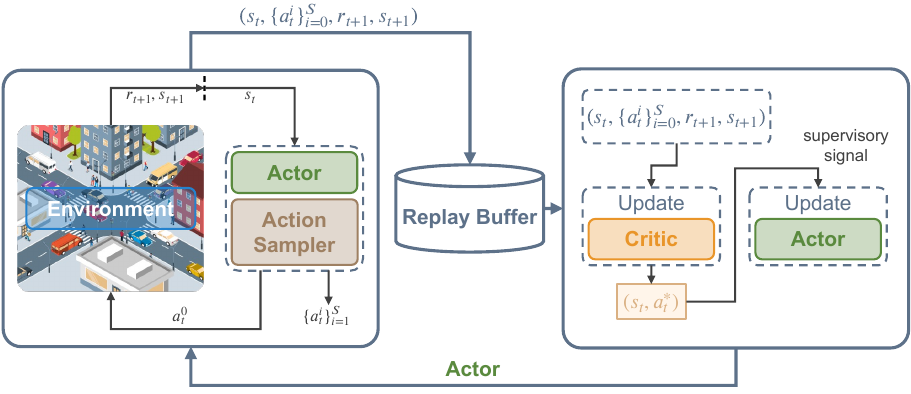}
\caption{Pipeline of the training process for the reinforced imitative trajectory planning}
\label{RITP}
\end{figure*}

As discussed in Section \ref{sec_introduction}, training a deep neural network-based policy for urban trajectory planning using conventional RL methods is challenging, as the gradients provided by RL are insufficient for optimizing the policy. Given that IL offers better convergence properties, a natural solution is to combine RL with IL, using RL to explore the state and trajectory spaces, while employing IL to optimize the policy. However, IL requires ground truth, meaning that when the RL agent interacts with the environment, IL must access the optimal future trajectory for each state, which is intractable. This represents a key challenge of this approach. Therefore, we propose the RITP method, which integrates IL into RL within an actor-critic framework, as shown in Fig. \ref{RITP}. RITP incorporates an action sampler in addition to the actor to generate potential future trajectories, estimates their critic values using the critic network, and selects the optimal trajectory based on these values. The optimal trajectory is then used as a supervisory signal of IL to update the actor network.

Specifically, at each time step $t$, the actor takes an action $a_t^0$ given a state $s_t$. At the same time, the action sampler generates $S$ actions for the same state $s_t$, i.e., $\{a_{t}^{i}\}_{i=1}^{S}$, where $i$ is the index of the action. The actions are defined as trajectories $\{a_{t}^{i} \mid a_{t}^{i}=\{\mathbf{p}_{t+j}^{i}\}_{j=1}^{T_\text{p}}\}_{i=0}^{S}$, with spatial location $\mathbf{p}_{t+j}^{i}=(x^i_{t+j}, y^i_{t+j})$ at $t+j$ time step, longitudinal and lateral coordinates $x^i_{t+j}, y^i_{t+j}$, and planning time horizon $T_\text{p}$. Afterward, only the action generated by the actor, i.e., $a_t^0$, is sent to the environment, which then outputs the reward $r_{t+1}$ and the subsequent state $s_{t+1}$. Next, observing the state $s_{t+1}$, the actor and action sampler output actions $\{a_{t+1}^{i}\}_{i=0}^{S}$, and the process repeats accordingly. 

While the actor and the action sampler interact with the environment, the experience, defined as a tuple $(s_t, \{a_t^i\}_{i=0}^{S}, r_{t+1}, s_{t+1} )$, is added to the replay buffer at each time step. Subsequently, a mini-batch of experiences is sampled from the replay buffer to update the critic network. Afterward, for each experience, the action with the highest state-action value estimated by the critic network is selected from $\{a_t^i\}_{i=0}^{S}$ as the optimal action $a_t^*$. Finally, the optimal state-action pair $(s_t, a_t^*)$ is used to update the actor via IL, and the updated actor is employed to interact with the environment at the next time step.

\subsection{Environment}\label{subsec_environment}
The environment can be built on a real-world urban automated driving dataset. During training, the environment is in non-reactive log-replay mode, where the other traffic participants, such as vehicles, pedestrians, and cyclists, do not react to the behavior of the ego agent and simply act according to the logged data. Upon receiving the action $a_{t}^{0}$ from the actor, the environment propagates the motion of the ego vehicle using an LQR-based trajectory tracker \cite{9517448} and the kinematic bicycle model. Note that an action is defined as a trajectory to be tracked by a downstream controller, rather than as multi-step control signals, primarily for the convenience of action sampling and post-processing. Additionally, the LQR-based trajectory tracker can be substituted with other controllers, such as a model predictive controller.

\subsection{State Representation}\label{subsec_state_representation}
The query-centric paradigm \cite{Zhou_2023_CVPR} is adopted to encode the traffic scene context. Consider a scenario with $A$ agents (including the ego agent and the other traffic participants),  $M$ map polygons (either from the upstream perception module or the high-definition map), and $T$ historical time steps. The $i$-th agent's information at historical time step $t-g$ ($g \in \{T-1, T-2,...,0\}$) consists of the spatial position $\mathbf{p}_{t-g}^{i}=(x_{t-g}^{i}, y_{t-g}^{i})$, the heading $h_{t-g}^i$, the temporal position ${t-g}$, the velocity $\mathbf{v}_{t-g}^i=(v_{{t-g},x}^i,v_{{t-g},y}^i)$, and the semantic attributes (e.g., agent type). Each map polygon (e.g., lanes, stop lines, and crosswalks) has $P$ sample points with spatial positions $\mathbf{p}^{i}=(x^{i}, y^{i})$, orientations $h^i$, and semantic attributes (e.g., traffic light status of a lane).

The query-centric paradigm builds an individual local spacetime coordinate system for the $i$-th agent at time step ${t-g}$ based on the position $\mathbf{p}_{t-g}^{i}$ and the heading $h_{t-g}^i$ or for the $i$-th map polygon based on the position and orientation of the first sample point of the polygon. Then, the information of an agent at time step ${t-g}$ is represented as $\mathbf{a}_{t-g}^i=(\Vert \mathbf{p}_{t-g}^{i} - \mathbf{p}_{t-g-1}^{i} \Vert_2, \langle \mathbf{h}_{t-g}^i, \mathbf{p}_{t-g}^{i}\rangle, \Vert \mathbf{v}_{t-g}^i \Vert_2, \langle \mathbf{h}_{t-g}^i, \mathbf{v}_{t-g}^i \rangle)$, where heading vector $\mathbf{h}_{t-g}^i=(\cos h_{t-g}^i, \sin h_{t-g}^i)$, and $\langle \cdot, \cdot \rangle$ denotes the angle between two vectors. The information of a sample point of a map polygon is represented as $\mathbf{m}^i=\Vert \mathbf{p}^{i} - \mathbf{p}^{i-1} \Vert_2$.

In order to incorporate the relative information between two spatial-temporal scene elements (e.g., an agent at some time step ${t-g}$ or a lane), a relative spatial-temporal positional encoding is employed. In particular, given a pair of spatial-temporal positions $(\mathbf{p}_{t-g}^i, h_{t-g}^i, {t-g})$ and $(\mathbf{p}_{t-g^\prime}^j, h_{t-g^\prime}^j, {t-g^\prime})$, the positional encoding is defined as $\mathbf{r}_{t-g^\prime \rightarrow t-g}^{j \rightarrow i} = (\Vert \mathbf{p}_{t-g^\prime}^{j} - \mathbf{p}_{t-g}^{i} \Vert_2, \text{atan2}(y_{t-g^\prime}^j - y_{t-g}^i, x_{t-g^\prime}^j - x_{t-g}^i) - h_{t-g}^i, h_{t-g^\prime}^j - h_{t-g}^i, g-g^\prime)$. Additionally, the time gap $g-g^\prime$ and the subscripts of the encodings for map polygons are omitted, i.e., $\mathbf{r}^{j \rightarrow i}$. 

\subsection{Actor Network: MotionFormer}\label{subsec_actor_network}
\begin{figure}[h]
\centering
\includegraphics[width=0.45\textwidth]{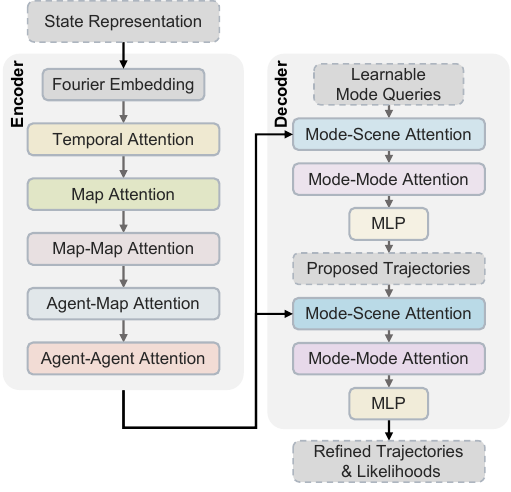}
\caption{Pipeline of MotionFormer}
\label{actor}
\end{figure}

Trajectory planning can be considered as a special case of trajectory prediction. Although trajectory prediction involves all agents while trajectory planning focuses solely on the ego agent, both tasks infer future trajectories based on traffic scene understanding. Therefore, a transformer-based actor network, referred to as MotionFormer, is constructed using QCNet architecture \cite{Zhou_2023_CVPR} for saving effort in actor network design, as shown in Fig. \ref{actor}.

MotionFormer comprises an encoder that generates scene encodings and a decoder that decodes $K$ future trajectories over a horizon of $T_\text{p}$ time steps for each agent along with the estimated likelihoods utilizing the scene encodings. Each future trajectory is called a mode. During training, the trajectory with the highest likelihood is considered as the ego agent's action. Additionally, the predicted trajectories for other agents are necessary for downstream post-optimization during testing, as explained in Section \ref{sec_post_optimizer}.

In the encoder, the Fourier embedding block first maps the state representation of the agents and map polygons, i.e., $\mathbf{a}_{t-g}^i$ and $\mathbf{m}^i$, to the Fourier features \cite{NEURIPS2020_55053683} to enhance the learning of high-frequency signals. These Fourier features are then combined with the semantic attributes, such as the traffic light status of a map polygon and an agent's type, and processed through a multi-layer perceptron (MLP) to produce agent embeddings of shape $[A,T,D]$ and map embeddings of shape $[M,P,D]$. Here, $P$ denotes the number of points per polygon, and $D$ denotes the hidden feature dimension. Also, the relative positional encodings are mapped to Fourier features and passed through an MLP to generate relative positional embeddings.

To extract the temporal information of the agents, the temporal attention block performs self-attention across the agents' historical time steps. Next, with the relative positional embeddings between the sample points of each polygon, the map attention block uses the map embedding for the first sample point to attend to those for the other sample points within the same polygon, reducing the map embeddings from $[M,P,D]$ to $[M,D]$. Afterward, the map-map, agent-map, and agent-agent attention blocks perform self-attention across $M$ polygons, cross-attention between agents and maps, and self-attention across $A$ agents, respectively, to capture interactions between scene elements. The outputs of these three blocks are collectively referred to as scene encodings, i.e., map encodings of shape $[M,D]$ and agent encodings of shape $[A,T,D]$.

The decoder concatenated with the encoder first uses learnable mode queries to attend to the scene encodings, followed by the mode-mode self-attention and MLP processing, producing the proposed trajectories. Then, the proposed trajectories are processed through additional refinement blocks, including a mode-scene attention block, a mode-mode attention block, and MLPs to generate the refined trajectories and the corresponding likelihoods. Each mode-scene attention block includes cross-attention operations between modes and agent encodings and between modes and map encodings, thereby updating the mode queries with the scene encodings. Each mode-mode attention block employs self-attention across $K$ modes to enhance the diversity of the modes.

\begin{figure}[h]
\centering
\includegraphics[width=0.45\textwidth]{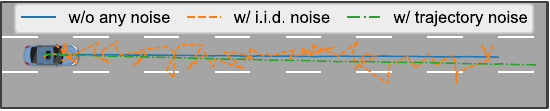}
\caption{Comparison between the trajectories without any noise, with i.i.d. noise, and with the proposed trajectory noise}
\label{trajectory noise}
\end{figure}

In RL, exploration noise for actions is necessary. Standard RL algorithms add independently and identically distributed (i.i.d.) noise to each dimension of an action. However, the i.i.d. noise reduces the smoothness of the action defined as a trajectory, as depicted by the orange dashed line in Fig. \ref{trajectory noise}. Therefore, we propose to add random trajectory noise to the refined trajectory with the highest estimated likelihood for the ego agent to facilitate exploration during training:
\begin{equation}
\begin{aligned}
	\tilde{a}_t &= \delta (a_t)\\
	&= \left\{\left(x_{t+j}+\alpha_j \epsilon \cos \theta_{\epsilon}, y_{t+j}+\alpha_j \epsilon \sin \theta_{\epsilon}\right)\right\}_{j=1}^{T_\text{p}} \\
	\epsilon &\sim \mathcal{N}\left(0, \beta \Vert \mathbf{p}_{t+T_\text{p}} - \mathbf{p}_{t}\Vert_2^2\right)\\
	\theta_\epsilon &\sim U[0,2\pi) \\
	\alpha_j &= j / T_\text{p}, \\
\end{aligned}
\label{def_trajectory_noise}
\end{equation}
where constant $\beta > 0$ controls the strength of exploration, $U$ denotes the uniform distribution, $\mathbf{p}_{t}$ denotes the ego agent's spatial position at time step $t$, and $\mathbf{p}_{t+T_\text{p}}$ denotes the final spatial position of the action. Intuitively, the magnitude of the noise added to each position on the trajectory increases linearly over planning time step $j$ until it reaches a maximum value, which is determined by the exploration strength and displacement of the trajectory, as depicted by the green dash-dot line in Fig. \ref{trajectory noise}.

\subsection{Action Sampler}\label{subsec_action_sampler}
Ideally, the actor should imitate the optimal action $a^*$ that maximizes the critic given a state $s$, i.e., $a^*=\arg \max_{a\in \mathcal{A}} Q(s,a)$. However, finding the optimal action is our goal in the first place. It is intractable because the action space $\mathcal{A}$ for AD is continuous, high-dimensional, and with kinematic and dynamic constraints. Therefore, TRAVL \cite{Zhang_Guo_2022} samples spline-based actions that are physically feasible for the ego agent in the Frenet frame of the road \cite{5509799}. Inspired by TRAVL, we propose sampling actions based on piecewise polynomials, allowing more flexible driving maneuvers.

\begin{figure}[h]
\centering
\includegraphics[width=0.45\textwidth]{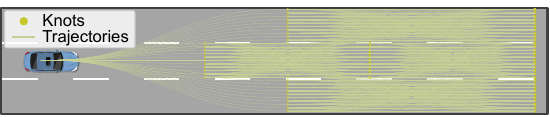}
\caption{Sampling actions based on piecewise polynomials}
\label{action samples}
\end{figure}

\begin{figure}[h]
\centering
\includegraphics[width=0.45\textwidth]{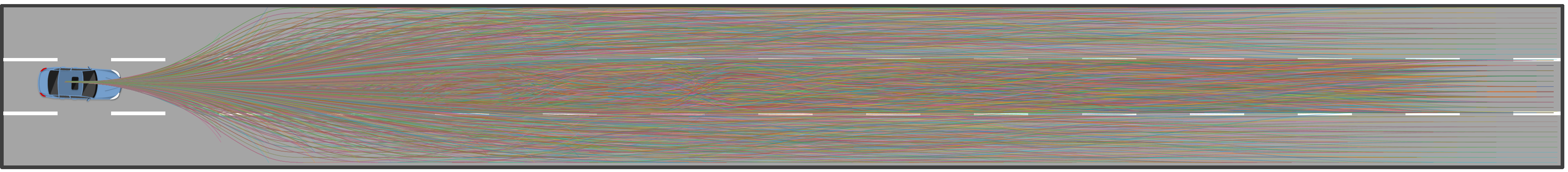}
\caption{An example of action samples (only 1\% of them are plotted for clarity) for a certain state}
\label{all action samples}
\end{figure}

Specifically, an action is decoupled into longitudinal and lateral trajectories
%\begin{equation}
%    \left\{
%    \begin{aligned}
%        x_i(t) &= p_{i,0}+p_{i,1} t+p_{i,2} t^{2}+p_{i,3} t^{3}+p_{i,4} t^{4} \\
%        y_i(t) &= q_{i,0}+q_{i,1} t+q_{i,2} t^{2}+q_{i,3} t^{3}+q_{i,4} t^{4} \\
%               &\quad +q_{i,5} t^{5}
%    \end{aligned}
%    \right.
%\end{equation}
\begin{equation}
    \begin{aligned}
        x_i(t) &= \boldsymbol{p}_{xi}^\top \cdot \left[1, t, t^2, t^3, t^4\right]^\top \\
        y_i(t) &= \boldsymbol{p}_{yi}^\top \cdot \left[1, t, t^2, t^3, t^4, t^5\right]^\top
    \end{aligned}
\end{equation}
where $x_i(t)$ and $y_i(t)$ denote the $i$-th piece longitudinal and lateral trajectories in the Frenet frame, respectively and $\boldsymbol{p}_{xi}=\left(p_{xi, 0}, p_{xi, 1}, \ldots, p_{xi, 4}\right)^\top$ and $\boldsymbol{p}_{yi}=\left(p_{yi, 0}, p_{yi, 1}, \ldots, p_{yi, 5}\right)^\top$ are vectors of coefficients, such that each piece of a trajectory is smoothly connected with its predecessors and successors at the knots. By sampling knots with different lateral positions for lane-keeping and lane-changing maneuvers, a set of trajectories (i.e., action samples) is obtained, as shown in Fig. \ref{action samples}, where the lane-changing and lane-keeping trajectories consist of two pieces and three pieces, respectively. In practice, denser samples (up to 230,000 trajectories per state in our experiments) can be generated to minimize the risk of missing the optimal action by adjusting the time length of each piece and varying the longitudinal and lateral positions of the knots, as shown in Fig. \ref{all action samples}.

\subsection{Critic Network: CriticFormer}\label{subsec_critic_network}
\begin{figure}[h]
\centering
\includegraphics[width=0.35\textwidth]{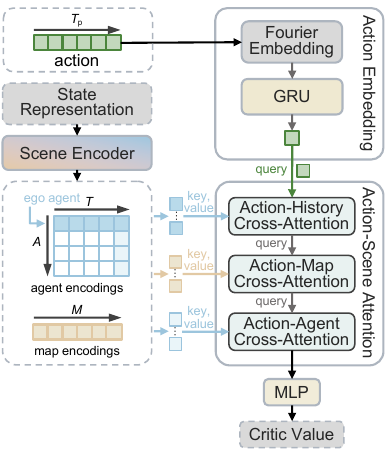}
\caption{Pipeline of CriticFormer}
\label{critic}
\end{figure}

In RL, the critic network evaluates how good an action is in a certain state. Aiming this, our critic network, namely CriticFormer, consists of a scene encoder, an action embedding block, and an action-scene attention block, as shown in Fig. \ref{critic}. Essentially, CriticFormer embeds the action into a query to attend to traffic scene encodings and then generates the critic value. To be specific, considering the similarity of the actor and critic networks in traffic scene understanding, CriticFormer employs a scene encoder with the same architecture as the actor network's encoder to encode the scene context. Furthermore, the action embedding block transforms the action into Fourier features. The Fourier features are further embedded by a GRU \cite{cho-etal-2014-learning}, the final hidden state of which serves as the action embedding.

In the action-scene attention block, the action embedding is successively updated by three cross-attention layers (action-history cross-attention, action-map cross-attention, and action-agent cross-attention) to attend to the ego agent's encodings across $T$ historical time steps, the map encodings, and the surrounding agents' encodings at the current time step, respectively. For the ego agent at the current spatial-temporal position $(\mathbf{p}_t^\text{ego}, h_t^\text{ego}, t)$ and the $i$-th agent or $i$-th map polygon, the positional encodings for the three cross-attention layers are given by $\mathbf{r}_{t-g \rightarrow t}^{\text{ego} \rightarrow \text{ego}}$, $\mathbf{r}^{i \rightarrow \text{ego}}$, and $\mathbf{r}_{t \rightarrow t}^{i \rightarrow \text{ego}}$. Finally, the critic value is obtained by passing the updated query through an MLP.

\subsection{Training Objectives}\label{subsec_training_objectives}
Following QCNet \cite{Zhou_2023_CVPR}, the ego agent's action is parameterized as a mixture of Laplace distributions:
\begin{equation}
	\begin{aligned}
	&f\left(\left\{\mathbf{p}_{t+j}\right\}_{j=1}^{T_{\text{p}}}\right)\\
	&=\sum_{k=1}^{K} \rho_{k} \prod_{j=1}^{T_{\text{p}}} \text {Laplace}\left(\mathbf{p}_{t+j} \mid \boldsymbol{\mu}_{t+j}^{k}, \boldsymbol{\sigma}_{t+j}^{k}\right)
	\end{aligned}
\end{equation}
with the mixing coefficient $\rho_{k}$, the location $\boldsymbol{\mu}_{t+j}^{k}$, and the scale $\boldsymbol{\sigma}_{t+j}^{k}$. MotionFormer is trained by imitation learning with a total loss defined as the sum of a classification loss $\mathcal{L}_{\text{c}}$, a regression loss for the proposed trajectory $\mathcal{L}_{\text{p}}$, and a regression loss for the refined trajectory $\mathcal{L}_{\text{r}}$:
\begin{equation}
	\mathcal{L}_{\text{actor}} = \mathcal{L}_{\text{c}} + \mathcal{L}_{\text{p}} + \mathcal{L}_{\text{r}},
	\label{actor_obj}
\end{equation}
where 
\begin{equation}
	\begin{aligned}
	\mathcal{L}_{\text{c}} &= -\log \sum_{k=1}^{K} \hat{\rho}_{k} \text {Laplace}\left(\mathbf{p}_{t+T_{\text{p}}}^{\text{target}} \mid \hat{\boldsymbol{\mu}}_{t+T_{\text{p}}}^{k}, \hat{\boldsymbol{\sigma}}_{t+T_{\text{p}}}^{k}\right)\\
	\mathcal{L}_{\text{p}} &= -\frac{1}{T_{\text{p}}} \log \prod_{j=1}^{T_{\text{p}}} \text {Laplace}\left(\mathbf{p}_{t+j}^{\text{target}} \mid \check{\boldsymbol{\mu}}_{t+j}^{*}, \check{\boldsymbol{\sigma}}_{t+j}^{*}\right)\\
	\mathcal{L}_{\text{r}} &= -\frac{1}{T_{\text{p}}} \log \prod_{j=1}^{T_{\text{p}}} \text {Laplace}\left(\mathbf{p}_{t+j}^{\text{target}} \mid \hat{\boldsymbol{\mu}}_{t+j}^{*}, \hat{\boldsymbol{\sigma}}_{t+j}^{*}\right),
	\end{aligned}
	\label{actor loss detail}
\end{equation}

In Eq. (\ref{actor loss detail}), $\{\mathbf{p}_{t+j}^{\text{target}}\}_{j=1}^{T_{\text{p}}}$ denotes the supervisory signal, i.e., $a_t^*=\arg \max_{a\in \{a_t^i\}_{i=0}^{S}} Q(s_t,a)$. Remind that $\{a_t^i\}_{i=0}^{S}$ is the set of the ego agent's actions output by the actor and action sampler. $\hat{\boldsymbol{\mu}}_{t+j}^{k}$ and $\hat{\boldsymbol{\sigma}}_{t+j}^{k}$ are the $j$-th location and scale of the $k$-th refined trajectory, whose likelihood is $\hat{\rho}_k$. When computing the gradient of $\mathcal{L}_{\text{c}}$, the gradients of the locations and scales are omitted, and only the gradients of the mixing coefficients are retained. $\{\check{\boldsymbol{\mu}}_{t+j}^{*}\}_j^{T_{\text{p}}}$ and $\{\check{\boldsymbol{\sigma}}_{t+j}^{*}\}_j^{T_{\text{p}}}$ denote the locations and scales of the best-proposed trajectory among the $K$ proposed trajectories in terms of the displacement error $\Vert \mathbf{p}_{t+T_{\text{p}}}^{\text{target}} - \check{\boldsymbol{\mu}}_{t+T_{\text{p}}}^{k} \Vert_2$. Similarly, $\{\hat{\boldsymbol{\mu}}_{t+j}^{*}\}_j^{T_{\text{p}}}$ and $\{\hat{\boldsymbol{\sigma}}_{t+j}^{*}\}_j^{T_{\text{p}}}$ denote the locations and scales of the best-refined trajectory. To facilitate stable training of the refinement blocks, the gradients of $\check{\boldsymbol{\mu}}_{t+j}^{*}$ and $\check{\boldsymbol{\sigma}}_{t+j}^{*}$ are omitted.

In practice, the MotionFormer is optimized to predict not only the ego agent's trajectory but also the trajectories of surrounding agents, whose supervisory signals are derived from the dataset. Because the predicted trajectories for surrounding agents are useful for the post-optimization (see Section \ref{sec_post_optimizer}). This training strategy is referred to as multi-agent supervision hereafter.

For CriticFormer, the noise in Eq. (\ref{q_obj}) is replaced with the clipped version of the trajectory noise defined as Eq. (\ref{def_trajectory_noise}), resulting in a target
\begin{equation}
	q =r+\gamma \min _{i=1,2} Q_{\theta_{i}^{\prime}}\left(s^{\prime}, \delta_c \left(\pi_{\phi^{\prime}}\left(s^{\prime}\right)\right)\right)
\end{equation}
Here, $\delta_c$ denote the same trajectory noise function as $\delta$ in Eq. (\ref{def_trajectory_noise}) except for 
\begin{equation}
\begin{aligned}
	\epsilon \sim \text{clip} & \left(\mathcal{N} \left(0, \beta^\prime \Vert \mathbf{p}_{t+T_\text{p}} - \mathbf{p}_{t}\Vert_2^2\right)\right.,\\
	 &\left. -c \Vert \mathbf{p}_{t+T_\text{p}} - \mathbf{p}_{t} \Vert_2^2,	 c \Vert \mathbf{p}_{t+T_\text{p}} - \mathbf{p}_{t} \Vert_2^2  \right) \\
\end{aligned}
\label{def_clipped_trajectory_noise}
\end{equation}
where $c$ is a constant that defines the boundaries of the clipping interval. In addition, $d$-delayed policy updates are employed to improve the quality of policy updates \cite{pmlr-v80-fujimoto18a}. RITP algorithm is summarized in Algorithm \ref{algo: RITP}.
\begin{algorithm}[!h]
\caption{Reinforced Imitative Trajectory Planning Algorithm}
\label{algo: RITP}
\LinesNumbered
Initialize actor network $\pi_\phi$ and critic networks $Q_{\theta_1}$, $Q_{\theta_2}$ with parameters $\phi$, $\theta_1$, $\theta_2$\;
Initialize target networks $\phi^\prime \leftarrow \phi$, $\theta_1^\prime \leftarrow \theta_1$, $\theta_2^\prime \leftarrow \theta_2$\;
Initialize replay buffer $\mathcal{B}$\;
Initialize $t=0$\;
\While{$t < $ maximum number of steps}{
	Select action with exploration noise in state $s$: $a^0 \leftarrow \delta(\pi_\phi(s))$, with $\delta$ defined by Eq. (\ref{def_trajectory_noise})\;
	Generate $S$ action samples $\{a^i\}_{i=1}^{S}$ using the action sampler\;
	Send $a^0$ to environment, receive reward $r^\prime$ and visit new state $s^\prime$\;
	Store experience $(s, \{a^i\}_{i=0}^{S}, r^\prime, s^\prime)$ in $\mathcal{B}$\;
	\BlankLine
	Sample mini-batch of $N$ experiences $(s, \{a^i\}_{i=0}^{S}, r^\prime, s^\prime)$ from $\mathcal{B}$\;
	Select action with clipped noise in sampled next state $s^\prime$: $a^\prime \leftarrow \delta_c (\pi_{\phi^\prime} (s^\prime))$, with $\delta_c$ defined by Eqs. (\ref{def_trajectory_noise}) and (\ref{def_clipped_trajectory_noise})\;
	Compute target critic value: $q \leftarrow r+\gamma \min _{i=1,2} Q_{\theta_{i}^{\prime}} \left( s^{\prime}, \delta_c \left( \pi_{\phi^{\prime}} \left( s^{\prime} \right)\right)\right)$\;
	Update critic networks to minimize $\qquad \qquad \qquad$ $\min _{\theta_i} N^{-1} \Sigma (q-Q_{\theta_i}(s,a))^2$\;
	\If{$t \mod d$ is $0$}{
		Find optimal action in state $s$: $\qquad \qquad \qquad$ $a^*=\arg \max_{i} Q_{\theta_1}(s,a^i)$\;
        Update actor network with target $a^*$ to minimize $N^{-1} \mathcal{L}_{\text{actor}}$ defined by Eqs. (\ref{actor_obj}) and (\ref{actor loss detail})\;
        Update target networks: $\qquad \qquad \qquad$ $\theta_i^\prime \leftarrow \xi \theta_i + (1-\xi) \theta_i^\prime$, $\qquad \qquad \qquad \qquad \qquad$ $\phi_i^\prime \leftarrow \xi \phi_i + (1-\xi) \phi_i^\prime$\;
	}
	$t \leftarrow t+1$\;
}
\end{algorithm}

\section{Bayesian RewardFormer}\label{sec_bayesian_rewardformer}
The reward function is critical in RL. However, few studies concentrate on learning effective reward functions for urban AD. Besides, as introduced in Section \ref{sec_introduction}, existing IRL methods for AD simplify the reward functions with the linear structure assumption, which limits the application of RL in urban AD. Addressing the reward function designing issue, we adapt our previous work, approximate variational reward learning (AVRL) \cite{AVRL}, to urban AD. Due to space constraints, the following content will not cover the detailed derivation for AVRL.

\subsection{Model Structure}
	\label{subsec_model_structure}
AVRL models a reward function as a deep Bayesian neural network (BNN) \cite{neal2012bayesian}, which estimates not only the reward for a state but also the uncertainty of the reward (i.e., how unconfident the model is). With variational inference \cite{blei2017variational}, the deep BNN can be approximated by a deep neural network with dropout layers \cite{hinton2012improving}. To encode the traffic scene context effectively, we also employ the query-centric state representation (see Section \ref{subsec_state_representation}) and propose a transformer-based reward function, namely Bayesian RewardFormer.

Bayesian RewardFormer, as shown in Fig. \ref{reward}, has the same architecture as the actor network's encoder except for two modifications. First, the temporal attention block is removed because Bayesian RewardFormer takes as input agents' current information $\mathbf{a}_{t}^i$ and map information $\mathbf{m}^i$, omitting agents' past information. Second, dropout layers are added before each fully connected layer in the attention blocks and the MLP to enable BNN approximation using dropout \cite{gal2016dropout}.
\begin{figure}[h]
\centering
\includegraphics[width=0.21\textwidth]{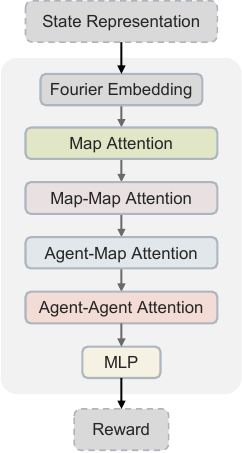}
\caption{Pipeline of Bayesian RewardFormer}
\label{reward}
\end{figure}
Specifically, let $\mathbf{M}_{i}$ and $\mathbf{b}_i$ be the weight matrix of dimensions $D_i \times D_{i-1}$ and the bias of dimensions $D_i$ for the $i$-th fully connected layer of the attention blocks and MLP. After adding dropout layers, each weight matrix-dropout layer pair can be considered as a whole:
\begin{equation}
	\begin{split}
		\label{q omega}
		&\mathbf{W}_{i}=\mathbf{M}_{i} \cdot \operatorname{diag}\left(\left[z_{i, j}\right]_{j=1}^{D_{i-1}}\right)\\
		&z_{i, j} \sim \operatorname{Bernoulli}\left(p_{i}\right),
	\end{split}
\end{equation}
with dropout probabilities $p_i$ for $i=1, \ldots, L$ and $j=1, \ldots, D_{i-1}$. $L$ denotes the number of the fully connected layers of the attention blocks and MLP. 

\subsection{Training Objectives}\label{subsec_reward_training_objectives}
Defining $\boldsymbol{\omega}=\{\mathbf{W}_i\}_{i=1}^{L}$ as the set of weight matrices for all $L$ layers, the loss with respect to $\{\mathbf{M}_{i}\}_{i=1}^{L}$ and $\{\mathbf{b}_{i}\}_{i=1}^{L}$ for training Bayesian RewardFormer is 
\begin{multline}
	-\sum_{(s_t, a_t)} \log \frac{b^{R(\zeta | \widehat{\boldsymbol{\omega}})}}{\sum_{\zeta^{\prime}} b^{R\left(\zeta^{\prime} | \widehat{\boldsymbol{\omega}}\right)}} \\
	+ \sum_{i=1}^{L}\left(\frac{p_{i}}{2}\left\|\mathbf{M}_{i}\right\|_{2}^{2}+\frac{1}{2}\left\|\mathbf{b}_{i}\right\|_{2}^{2}\right),
	\label{reward_objective}
\end{multline}
with $\widehat{\boldsymbol{\omega}}$ sampled from the distribution of $\boldsymbol{\omega}$. In Eq. (\ref{reward_objective}), $a_t$ denotes the ego agent's action (trajectory) in state $s_t$ in the dataset. $\zeta = \{\tilde{s}_{t+j}\}_{j=1}^{T_\text{p}}$ denotes the set of the predicted states that the ego agent will visit after taking action $a_t$, assuming that the ego agent follows the kinematic bicycle model and the other agents are in a non-reactive log-replay mode. $\zeta^\prime$ represents the set of the predicted states for another possible action sampled based on polynomials in the Frenet frame of the road. $R(\zeta | \widehat{\boldsymbol{\omega}})$ is defined as $\sum_{j=1}^{T_\text{p}}R(\tilde{s}_{t+j} | \widehat{\boldsymbol{\omega}})$. The base $b$ of the exponentiation is set to 1.1, which differs from the natural constant $e$ employed in AVRL and improves the convergence when training the reward function.

\subsection{Predicted Uncertainty-Penalized Reward}\label{subsec_pup_reward}
Neural networks make arbitrary predictions outside the domain they were trained on, known as epistemic uncertainty \cite{gal2016uncertainty}. In the case of RL, a neural network-based reward function may output wrong rewards when the states differ from those in the dataset, thus leading to unexpected or even dangerous actions of the learned policy. Therefore, the reward at each RL time step is replaced with the predicted uncertainty-penalized reward \cite{AVRL}:
\begin{equation}
	\begin{split}
		\label{reward}
		r_t&=\sum_{j=1}^{T_\text{p}} \tilde{r}_{t+j}-\lambda \tilde{u}_{t+j} \\
		\tilde{r}_{t+j}&=\frac{1}{O} \sum_{o=1}^{O} R\left(\tilde{s}_{t+j} | \widehat{\boldsymbol{\omega}}_{o}\right) \\
		\tilde{u}_{t+j}&=\tau^{-1}\sigma^2 + \frac{1}{O} \sum_{o=1}^{O} \left[ R\left(\tilde{s}_{t+j} | \widehat{\boldsymbol{\omega}}_{o}\right)\right] ^2 -\tilde{r}_{t+j}^{2},
	\end{split}
\end{equation}
where $\tilde{s}_{t+j}$, $\tilde{r}_{t+j}$, and $\tilde{u}_{t+j}$ denote the predicted state, reward mean, and reward uncertainty at future time step $t+j$. The uncertainty consists of the aleatoric uncertainty \cite{gal2016uncertainty} ($\tau^{-1}\sigma^2$, with some model precision hyperparameter $\tau=1$ and $\sigma=1$) and the epistemic uncertainty (the reset terms except $\tau^{-1}\sigma^2$). The epistemic uncertainty is estimated by the variance of the $O$ reward samples generated by $O$ stochastic forward passes through Bayesian RewardFormer. Hyperparameter $\lambda$ controls the strength of uncertainty-penalization. With the reward defined in Eq. (\ref{reward}), the ego agent is encouraged to minimize the epistemic uncertainty while maximizing the reward, which helps the learned policy to avoid unfamiliar situations \cite{AVRL}.

\section{Post-Optimizer}\label{sec_post_optimizer}
The epistemic uncertainty is inevitable for neural networks, including MotionFormer. MotionFormer may output dangerous actions in unfamiliar situations, e.g., trajectories leading to pedestrian collisions. Therefore, a model-driven post-optimizer is developed to further refine the action outputted by MotionFormer in the testing stage.

The post-optimizer comprises trajectory proposal generation, QP path planning, and QP speed planning, all performed within the road's Frenet frame. In the trajectory proposal generation step, the post-optimizer processes the action $a$ outputted by MotionFormer into a trajectory proposal. Specifically, candidate trajectories (up to 1,500 trajectories in our experiments) are generated based on piecewise polynomials as shown in Fig. \ref{action samples}. Subsequently, a closed-loop simulation is conducted for these candidate trajectories using an LQR trajectory tracker and a kinematic bicycle model to obtain simulated rollouts, following the method described in Ref. \cite{pmlr-v229-dauner23a}. Each rollout is then evaluated based on criteria such as driving safety, traffic rule compliance, and progress along route, utilizing a rule-based evaluation framework established  in Ref. \cite{pmlr-v229-dauner23a}. This evaluation assigns each candidate trajectory a rule-based score. The final score for each candidate trajectory is calculated by adding the rule-based score and a preference score, which is defined as the L2 distance between the action $a$ and the candidate trajectory. Finally, the trajectory proposal is selected as the candidate trajectory with the highest final score. The motivation for this step is to preliminarily refine potentially dangerous actions outputted by MotionFormer, which could otherwise mislead the subsequent QP path and speed planning.

However, the generated trajectory proposals may be inflexible in handling diverse situations due to their predefined shapes. Therefore, QP path and speed planning are necessary to further refine the trajectory proposal. Specifically, a trajectory proposal is decoupled into a path profile $y(x)$ and a speed profile $x(t)$, represented as smoothing spline functions. Here, $x$, $y$, and $t$ represent the longitudinal coordinate, lateral coordinate, and time, respectively. Specifically, $y(x)$ is defined as
\begin{equation}
\label{smoothing_spline}
y(x)=\left\{\begin{array}{ll}
	y_{0}\left(x-x_{0}\right) & x \in\left[x_{0}, x_{1}\right) \\
	y_{1}\left(x-x_{1}\right) & x \in\left[x_{1}, x_{2}\right) \\
	\ldots & \ldots \\
	y_{n-1}\left(x-x_{n-1}\right) & x \in\left[x_{n-1}, x_{n}\right]
\end{array}\right.
\end{equation}
where $y_{i}(x)=p_{i, 0}+p_{i, 1} x+\ldots + p_{i, m} x^{m}$, $i=0, 1, ..., n-1$, is a $m$-th degree polynomial function with the coefficient $\boldsymbol{p}_{i}=\left(p_{i, 0}, p_{i, 1}, \ldots, p_{i, m}\right)^\top$, and $\{x_0, x_1, ..., x_n\}$ are the spline knots. The parameter vector of the smoothing spline function is denoted as $\boldsymbol{p} = \left(\boldsymbol{p}_{0}^\top, \boldsymbol{p}_{1}^\top, \ldots, \boldsymbol{p}_{n-1}^\top\right)^\top$. $x(t)$ has the same form as $y(x)$.

The QP path planning is mathematically described as
\begin{equation}
	\min _{y \in \Omega} J(y)=\sum_{i=0}^{3} w_{i} J_{i}(y)
	\label{qp_path_problem}
\end{equation}
subject to linear constraints $\boldsymbol{L}(y(x)) \le \boldsymbol{0}$, where $
\Omega=\left\{ \right. y:[x_0, x_n] \rightarrow \mathbb{R} \mid y, y^{(1)}, y^{(2)}, ..., y^{(m)}$
is absolutely continuous and  $\int_{x_0}^{x_n}\left(y^{(i)}\right)^{2} d x<\infty,$ $i=0, 1, ..., m\left.\right\}$, and $w_i$ are weights. $y^{(i)}$ denotes the $i$-th derivative of $y$. In detail, the objective functionals are defined as
\begin{equation}
\begin{array}{l}
	J_{0}(y)=\int_{x_0}^{x_n}(y(x)-\hat{y}(x))^{2} d x \\
	J_{1}(y)=\int_{x_0}^{x_n}\left(y^{(1)}(x)\right)^{2} d x \\
	J_{2}(y)=\int_{x_0}^{x_n}\left(y^{(2)}(x)\right)^{2} d x \\
	J_{3}(y)=\int_{x_0}^{x_n}\left(y^{(3)}(x)\right)^{2} d x,
\end{array}
\end{equation}
where $\hat{y}(x)$ represents the target path profile derived from the most likely trajectory outputted by MotionFormer. The linear constraints are related to collision avoidance, vehicle dynamics, and traffic regulations. For example, a constraint can be established based on the predicted trajectories of surrounding agents to avoid collisions.

Analogously, the QP speed planning is performed in the same manner as the QP path planning, except that the notation $x$ is replaced by $t$, $y$ is replaced by $x$, and $J_1$, which encourages low speed, is removed. The post-optimized action is then obtained by combining the path and speed profiles. Essentially, solving the QP problems for path and speed planning involves finding a smooth trajectory that remains close to the original prediction while explicitly considering collision avoidance, vehicle dynamics, traffic regulations, and other relevant factors. In addition, if the QP path or speed planning fails, the planned path profile will be the centerline of the lane, and the planned speed profile will be computed by an intelligent driver model (IDM \cite{PhysRevE.62.1805}) with the same settings as Ref. \cite{pmlr-v229-dauner23a}.

\section{Experiments}\label{sec_experiments}

\subsection{Experimental Settings}\label{subsec_experimental_settings}
\subsubsection{Dataset}\label{subsubsec_dataset}
We conducted experiments on the large-scale real-world automated driving nuPlan dataset \cite{caesar2021nuplan, karnchanachari2024towards} to validate the effectiveness of our approach. The nuPlan dataset consists of 1282 hours of diverse urban driving scenarios from four cities (Las Vegas, Boston, Pittsburgh, and Singapore), including merges, lane changes, protected or unprotected turns, interactions with pedestrians and cyclists, and stop control in intersections. Each scenario lasts approximately 15 seconds and records the geometric and semantic information of surrounding map elements and traffic participants at a sampling rate of 10 Hz. Additionally, the motion states of traffic participants are recorded. The traffic light statuses are encoded into map elements. Our experiments randomly selected 10,000 scenarios comprising 70.1\% training scenarios, 12.5\% validation scenarios, and 17.4\% testing scenarios.

\subsubsection{Simulator and Metrics}\label{subsubsec_simulator_and_metrics}
Our experiments utilized the simulation and evaluation framework provided by the nuPlan dataset which enables closed-loop simulation considering the interactions between traffic participants. The simulator receives the planned trajectory of 8 seconds from the planner, uses an LQR as the tracker, and updates the motion state of the ego agent based on the kinematic bicycle model. In our experiments, the surrounding agents were successively in a closed-loop non-reactive log-replay mode and an IDM-based reactive mode. In the IDM-based reactive mode, the surrounding agents followed the paths in the dataset, while IDMs controlled the longitudinal motion.

To evaluate the performance of a planner, the closed-loop non-reactive score and the closed-loop reactive score were adopted following the standard evaluation protocol of the nuPlan dataset. Each score is a weighted average of metrics, including no at-fault collisions, drivable area compliance, making progress, driving direction compliance, time-to-collision within bound, progress along route ratio, speed limit compliance, and comfort. Additionally, we defined a metric of success rate as the number of scenarios without at-fault collisions divided by the total number of scenarios.

\subsection{Baselines}\label{subsec_baselines}
Urban Driver \cite{UrbanDriver}, QCNet \cite{Zhou_2023_CVPR}, and PLUTO \cite{cheng2024pluto} were adopted as the baselines in our experiments. The Urban Driver model embeds vectorized state representation of agents and map polygons into local features which are then passed through a global attention layer to predict trajectories. The Urban Driver model was trained using open-loop imitation learning implemented by nuPlan. 

QCNet \cite{Zhou_2023_CVPR}, which won the championship of the Argoverse 2 multi-agent motion forecasting competition, processes query-centric state representation into scene encodings using a factorized attention mechanism and predicts multimodal trajectories with DETR-like decoders \cite{DETR}. We implemented QCNet in the nuPlan framework and trained it using open-loop imitation learning. Furthermore, the size of QCNet was reduced to enable training on a single NVIDIA RTX 4090 GPU with 24GB of VRAM, where the Urban Driver model and our MotionFormer, CriticFormer, and Bayesian RewardFormer were also trained. In order to compensate for the performance loss caused by model size reduction, a masked training strategy \cite{lan2024sept} and a self-distillation strategy \cite{NIPS2017_e1e32e23, Shen_2022_CVPR} were used for training QCNet, resulting in a variant of QCNet namely query-centric masked autoencoder (QCMAE). 

PLUTO \cite{cheng2024pluto} is a state-of-the-art (SOTA) planning method on the nuPlan dataset. It uses a longitudinal-lateral aware transformer with factorized self-attention to jointly model maneuvers and generate diverse, reference-line-conditioned trajectory proposals for flexible and safe driving. Furthermore, PLUTO employs contrastive imitation learning with novel data augmentations (e.g., state perturbation, agent dropout) to mitigate causal confusion and distribution shift commonly faced by traditional IL methods. Additionally, PLUTO integrates a rule-based safety verification module to improve driving safety. We trained PLUTO using the hyperparameters specified in Ref. \cite{cheng2024pluto}, on the same dataset and hardware setup as our RITP model and other baseline methods.

\subsection{Implementation details}
We first trained Bayesian RewardFormer using AVRL. Then, QCMAE was trained using the masked training and self-distillation strategies. Next, with the trained QCMAE as a warm start, MotionFormer was trained using the proposed RITP method. The optimizer, learning rate scheduler, and hyperparameters used for training are summarised in Table~\ref{tab:details}. For more details, please refer to the source code at \url{https://github.com/Zigned/nuplan_zigned}.
\begin{table*}[h]
\centering
\begin{threeparttable}
\caption{Training configuration}\label{tab:details}
\begin{tabular}{@{}lccc@{}}
\toprule
Hyperparameters & AVRL  & QCMAE & (H)RITP\\
\midrule
Dropout probability $p_i$    & 0.1 & 0.1  & 0.1  \\
Planning horizon $T_{\text{p}}$    & 80   & 80  & 80  \\
Historical time horizon $T$    & -   & 50  & 50  \\
Hidden feature dimension $D$    & 64   & 64  & 64  \\
Number of modes $K$    & -   & 6  & 6  \\
Exploration strength $\beta$    & -   & -  & 0.1  \\
Policy noise strength $\beta$    & -   & -  & 0.2  \\
Clipping boundary $c$    & -   & -  & 0.5  \\
Optimizer    & Adam \cite{adam}   & AdamW \cite{adamw}  & AdamW  \\
Learning rate    & 3e-7 & 5e-5$\rightarrow$5e-4$\rightarrow$5e-5  & 5e-5  \\
Learning rate scheduler & - & OneCycle \cite{onecyclelr} & - \\
Batch size $N$    & 1   & 4  & 4, 1\tnote{1}   \\
Discount factor $\gamma$    & -   & -  & 0.99\\
Target update rate $\xi$    & -   & -  & 0.005\\
Update delay $d$    & -   & -  & 2\\
Uncertainty-penalization strength $\lambda$ & -   & -  & 1.5\\
Number of stochastic forward passes $O$ & 10 & - & - \\
Max number of experiences in replay buffer   & -   & -  & 1e4  \\
Total training epochs   & 10   & 60  & -  \\
Total training steps   & -   & -  & 1e5  \\
\botrule
\end{tabular}
\begin{tablenotes}
\footnotesize
\item[1] 4 for collecting experiences and 1 for optimization.
\end{tablenotes}
\end{threeparttable}
\end{table*}

\subsection{Comparison with Baselines}\label{subsec_comparision_with_baselines}
The results of the baselines and proposed approach on the testing set are summarized in Table~\ref{tab: comparison with baselines}. Comparing the results, it can be concluded that our RITP method substantially outperforms the Urban Driver and QCMAE in terms of both the closed-loop non-reactive score (over 125.0\% improvement) and the closed-loop reactive score (over 113.1\% improvement). This remarkable improvement over QCMAE is particularly noteworthy given that RITP shares the same policy network structure, highlighting the effectiveness of the proposed method. Furthermore, RITP achieved competitive composite scores compared to that of the SOTA method PLUTO, even outperforming PLUTO on safety-related metrics (NAFC, TTCWB, and, SR). The slight decrease in scores on metrics related to traffic rules and driving tasks (DAC, DDC, and PARR) is deemed acceptable given the paramount importance of safety in automated driving. Regarding the comfort metric CMFT, RITP and PLUTO are both inferior to UrbanDriver and QCMAE. This is a consequence of compromises made in their design to handle challenging emergency situations (e.g., hard braking). Note that fewer collisions happened in the closed-loop reactive mode than in the closed-loop non-reactive mode because the surrounding agents in the latter mode automatically controlled the distance to the ego agent to avoid collisions.

\begin{table*}[htbp]
  \centering
  \begin{threeparttable}
	\caption{Quantitative results on the testing set}
	\setlength{\tabcolsep}{4pt}  % 设置列间距
	\label{tab: comparison with baselines}
    \begin{tabular}{@{}l|c|cccccccc|cc@{}}
      \toprule
      Method & Mode & NAFC & TTCWB & DAC & DDC & SLC & MP & PARR & CMFT & SR & score  \\
      \midrule
      Urban Driver\cite{UrbanDriver} &\multirow{4}{*}{CL-NR} & 0.53 & 0.48 & 0.50 & 0.69 & 0.92 & 0.95 & 0.80 & \underline{0.97} & 0.53 & 0.32\\
      QCMAE && 0.54 & 0.49  & 0.68  & 0.95 & 0.98 & 0.96 & 0.76 & \textbf{0.99} & 0.55 & 0.36\\
      RITP (ours) && \textbf{0.94} & \textbf{0.90}  & \underline{0.97}  & \underline{0.99} & \textbf{1.00} & \underline{0.96} & \underline{0.83} & 0.81 & \textbf{0.94} & \underline{0.81}\\
      PLUTO \cite{cheng2024pluto} && \underline{0.93} & \underline{0.86} & \textbf{0.99}  & \textbf{0.99} & \underline{0.98} & \textbf{0.98} & \textbf{0.91} & 0.79 & \underline{0.93} & \textbf{0.84}\\
      \midrule
      Urban Driver\cite{UrbanDriver} &\multirow{4}{*}{CL-R} & 0.56 & 0.50 & 0.51 & 0.69 & 0.92 & 0.96 & \underline{0.82} & \underline{0.97} & 0.58 & 0.29\\
      QCMAE && 0.63 & 0.57  & 0.67  & 0.94 & 0.98 & 0.96 & 0.75 & \textbf{0.99} & 0.66 & 0.38\\
      RITP (ours) && \textbf{0.95} & \textbf{0.89}  & \underline{0.97}  & \underline{0.99} & \textbf{1.00} & \textbf{0.98} & 0.80 & 0.82 & \textbf{0.95} & \underline{0.81}\\
      PLUTO \cite{cheng2024pluto} && \underline{0.94} & \underline{0.88} & \textbf{1.00}  & \textbf{0.99} & \underline{0.98} & \underline{0.97} & \textbf{0.85} & 0.77 & \underline{0.95} & \textbf{0.83}\\
      \bottomrule
    \end{tabular}
    \begin{tablenotes}
	\footnotesize
	\item[] CL-NR: closed-loop non-reactive. CL-R: closed-loop reactive. NAFC: no at-fault collisions. TTCWB: time-to-collision within bound. DAC: drivable area compliance. DDC: driving direction compliance. SLC: speed limit compliance. MP: making progress. PARR: progress along route ratio. CMFT: comfort. SR: success rate. Higher is better for all metrics. The best result for each metric is in \textbf{bold}, and the second best result is \underline{underlined}.
	\end{tablenotes}
  \end{threeparttable}
\end{table*}

\subsection{Qualitative Results}\label{subsec_qualitative_results}
\begin{figure*}[!h]
\centering
\includegraphics[width=1.0\textwidth]{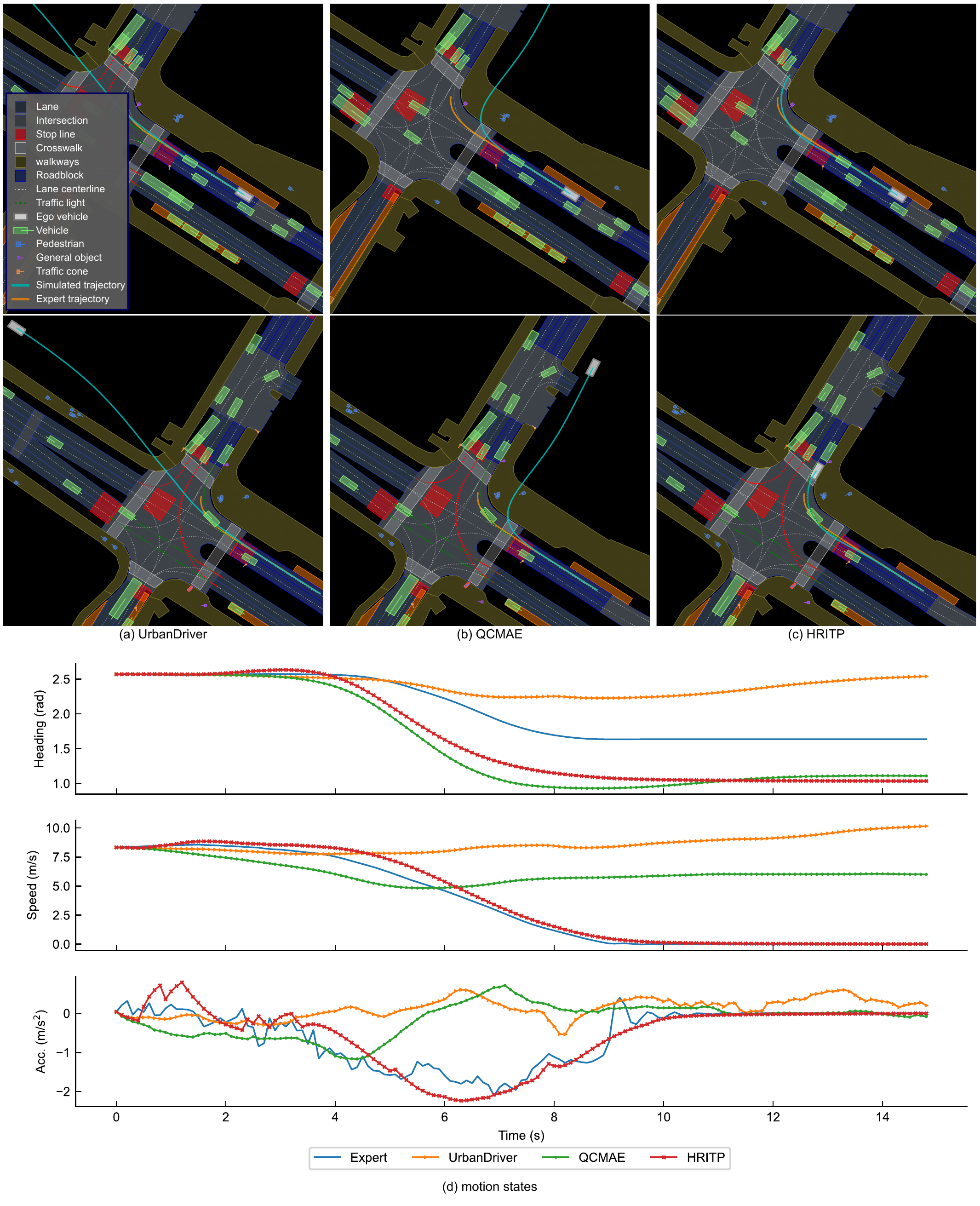}
\caption{A scenario of traversing an intersection}
\label{eee5}
\end{figure*}

This section presents some qualitative results on the testing set. Fig. \ref{eee5} illustrates a scenario involving traversing an intersection. Fig. \ref{eee5}(a)-(d) depict the expert's right turn trajectory (orange curves) and the corresponding ego vehicle trajectories generated by UrbanDriver, QCMAE, RITP, and PLUTO (blue curves). In these subfigures, the top and bottom halves represent the first and last frames of the simulation, respectively. The motion states (e.g., speed, acceleration) of the ego vehicle for each method are presented in Fig. \ref{eee5}(e). As shown in Fig. \ref{eee5}, UrbanDriver and QCMAE were able to drive within the lane at the beginning of the simulation. However, they subsequently departed from the drivable area, which was caused by covariate shift. In contrast, the ego vehicle controlled by RITP followed the preceding vehicle into the intersection and stopped appropriately when the preceding vehicle halted (possibly for pick-up or drop-off), effectively mimicking the expert's behavior. PLUTO also succeeded in navigating this scenario, although it exhibited less stable acceleration, as evident in Fig. \ref{eee5}(e).

\begin{figure*}[!h]
\centering
\includegraphics[width=1.0\textwidth]{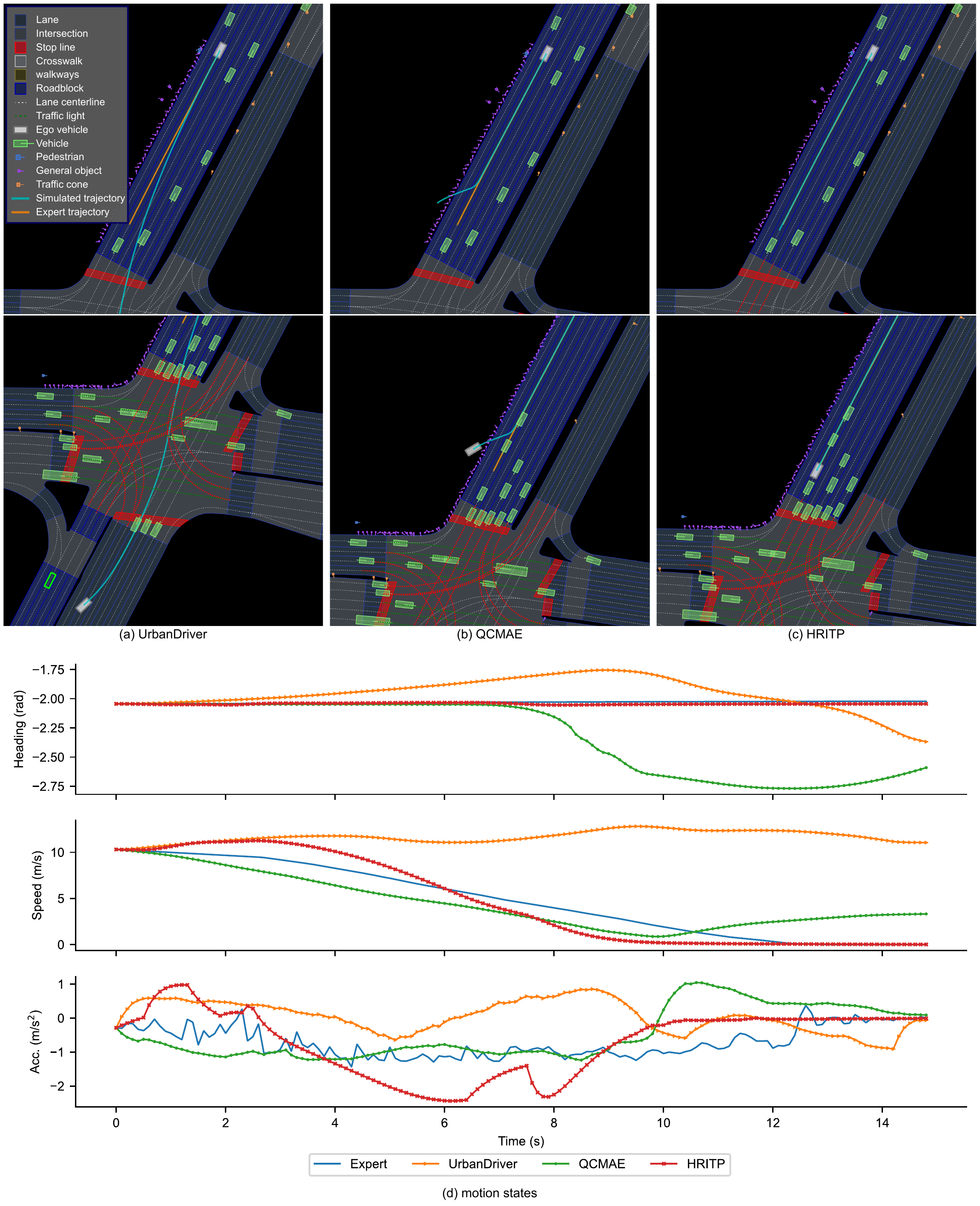}
\caption{A scenario of stopping near an intersection}
\label{207e}
\end{figure*}

Fig. \ref{207e} illustrates another scenario, where the ego vehicle was supposed to stop near the intersection. However, as depicted in Fig. \ref{207e}(a), the UrbanDriver quickly veered off, running a red light and even driving into the wrong lane. QCMAE performed significantly better during the initial 7 seconds but subsequently drove out of the drivable area over the following 8 seconds, as shown in Fig. \ref{207e}(b). In contrast, RITP successfully guided the ego vehicle to slow down and maintain a safe distance from the preceding vehicle, while PLUTO planned reasonable lane change maneuvers, as shown in Fig. \ref{207e}(c) and (d).

\subsection{Ablation Study}\label{subsec_ablation_study}
Closed-loop simulations were conducted on the validation set, covering both non-reactive and reactive modes, to analyze the impact of the following factors on model performance: the multi-step reasoning (MSR), the warm start (WS), the multi-agent supervision (MAS, see Section \ref{subsec_training_objectives}), and the hybrid-driven planning framework (Hybrid, see Section \ref{sec_post_optimizer}). Specifically, MSR refers to training MotionFormer to plan an 8-second trajectory (80 future positions), whereas without MSR, it was trained for a 1-second trajectory (2 future positions). WS indicates that MotionFormer was initialized with weights from the trained QCMAE model, whereas without WS, it was trained from scratch. Note that the standard TD3 \cite{pmlr-v80-fujimoto18a} algorithm was employed to train MotionFormer without MSR, while our RITP method was used in the other ablation experiments.

Table \ref{tab: ablation study} summarizes the results of the ablation study. As shown in Table \ref{tab: ablation study}, $\mathcal{M}_0$, which omits MSR, WS, MAS, and Hybrid planning, exhibits the poorest performance. Although $\mathcal{M}_1$ incorporated MSR, it also has poor closed-loop planning performance. The performance of $\mathcal{M}_2$ highlights the necessity of warm-starting the training with the trained QCMAE model. Furthermore, incorporating MAS ($\mathcal{M}_3$) results in a significant improvement in closed-loop non-reactive and reactive scores (39.4\% and 45.7\%, respectively) compared to approaches that solely optimize the ego-vehicle trajectory outputted by MotionFormer. This significant improvement is attributed to MAS simultaneously optimizing the predicted trajectories of all surrounding vehicles, thereby increasing data richness and facilitating better trajectory modeling. Moreover, with the hybrid planning framework, $\mathcal{M}_4$ further significantly enhances planning performance across most metrics compared to $\mathcal{M}_3$, with an acceptable reduction observed in CMFT. Specifically, the closed-loop non-reactive and reactive scores of $\mathcal{M}_4$ improved by 82.6\% and 62.7\% compared to $\mathcal{M}_3$.

\begin{table*}[htbp]
  \centering
  \begin{threeparttable}
    \caption{Ablation study of the impact of each factor on model performance on the validation set}
    \label{tab: ablation study}
	\setlength{\tabcolsep}{3pt}  % 设置列间距
    \begin{tabular}{@{}c|c|cccc|cccccccc|cc@{}}
      \toprule
      Mode & Model & MSR & WS & MAS & Hybrid & NAFC & TTCWB & DAC & DDC & SLC & MP & PARR & CMFT & SR & score  \\
      \midrule
      \multirow{5}{*}{CL-NR} & $\mathcal{M}_0$ & & & & & 0.30 & 0.26 & 0.06 & 0.95 & 0.99 & 0.47 & 0.40 & 0.60 & 0.31 & 0.02\\
       & $\mathcal{M}_1$& $\checkmark$ & & & & 0.31 & 0.29 & 0.27 & 0.89 & 0.93 & 0.88 & 0.74 & 0.70 & 0.32 & 0.13 \\
       & $\mathcal{M}_2$& $\checkmark$ & $\checkmark$ & & & 0.70 & 0.62 & 0.79 & 0.99 & 0.99 & 0.61 & 0.46 & 0.80 & 0.71 & 0.33 \\
       & $\mathcal{M}_3$& $\checkmark$ & $\checkmark$ & $\checkmark$ & & 0.76 & 0.72 & 0.78 & 0.99 & 0.99 & 0.78 & 0.59 & \textbf{0.97} & 0.77 & 0.46 \\
       & $\mathcal{M}_4$& $\checkmark$ & $\checkmark$ & $\checkmark$ & $\checkmark$ & \textbf{0.96} & \textbf{0.91} & \textbf{0.97} & \textbf{1.00} & \textbf{1.00} & \textbf{0.96} & \textbf{0.85} & 0.84 & \textbf{0.96} & \textbf{0.84} \\
      \midrule
      \multirow{5}{*}{CL-R} & $\mathcal{M}_0$ & & & & & 0.37 & 0.33 & 0.08 & 0.95 & 0.99 & 0.47 & 0.40 & 0.61 & 0.38 & 0.04\\
       & $\mathcal{M}_1$ & $\checkmark$ & & & & 0.38 & 0.32 & 0.30 & 0.88 & 0.93 & 0.88 & 0.74 & 0.70 & 0.39 & 0.14 \\
       & $\mathcal{M}_2$& $\checkmark$ & $\checkmark$ & & & 0.76 & 0.71 & 0.78 & 0.99 & 0.99 & 0.60 & 0.46 & 0.78 & 0.77 & 0.35 \\
       & $\mathcal{M}_3$& $\checkmark$ & $\checkmark$ & $\checkmark$ & & 0.84 & 0.80 & 0.80 & 0.99 & 0.99 & 0.77 & 0.57 & \textbf{0.98} & 0.84 & 0.51 \\
       & $\mathcal{M}_4$& $\checkmark$ & $\checkmark$ & $\checkmark$ & $\checkmark$ & \textbf{0.96} & \textbf{0.92} & \textbf{0.98} & \textbf{1.00} & \textbf{1.00} & \textbf{0.97} & \textbf{0.81} & 0.84 & \textbf{0.97} & \textbf{0.83} \\
      \bottomrule
    \end{tabular}
    % \begin{tablenotes}
    %   \item[a] This is a note.
    %   \item[b] Another note.
    % \end{tablenotes}
  \end{threeparttable}
\end{table*}

We also attempted to train $\mathcal{M}_1$and $\mathcal{M}_2$ using the standard TD3 algorithm instead of our RITP method. However, the training failed to converge, and the resulting model yielded erratic trajectories consisting of a set of disorganized positions.

\subsection{Runtime Analysis}\label{subsec_ablation_study}
This section provides an analysis of the computational time complexity and runtime of the proposed method. During inference, the computational time complexity of RITP stems primarily from the attention modules within MotionFormer and the post-optimizer components: trajectory proposal generation, QP path planning, and QP speed planning. Given the parameters $A$ (number of agents), $T$ (number of historical time steps), and $M$ (number of map polygons), the time complexity of the attention mechanisms dominates, specifically: temporal attention ($O(AT^2)$), map-map attention ($O(M^2)$), agent-map attention ($O(ATM)$), and agent-agent attention ($O(A^2T)$). The post-optimizer involves curve interpolation, LQR, and QP. These sub-problems can be solved efficiently using mature mathematical tools and have been proven to meet the real-time requirements of automated vehicles, as demonstrated by their use in systems like Baidu Apollo \cite{fan2018baidu}.

Table~\ref{tab: inference_time} presents the model inference time (mean plus/minus one standard deviation) in a dense traffic scenario involving 71 agents and 240 map polygons. The model was run on a single NVIDIA RTX 4090 GPU and an Intel Xeon 4310 processor (up to 3.3 GHz). The results indicate that the MotionFormer component exhibits very low latency. This efficiency is primarily attributable to the highly parallel CUDA architecture and dedicated Tensor Cores of the NVIDIA GPU, along with its efficient support for low-precision inference and batch computation. In contrast, the Post-Optimizer component incurs higher runtime. This is because the Post-Optimizer was implemented in Python, an interpreted language known for significantly lower execution efficiency compared to C++. Based on this, it is reasonable to expect that reimplementing the Post-Optimizer in C++ would yield significant computational efficiency improvements.

\begin{table}[htbp] % htbp 是表格位置的建议参数：here, top, bottom, page
  \centering % 表格居中
  \caption{Model inference time on the validation set} % 表格标题
  \label{tab: inference_time} % 用于交叉引用的标签
  \begin{tabular}{l|c} % 定义列数和对齐方式 (l:左对齐, c:居中, r:右对齐)
    \toprule % 顶部线条
    Component        & Online Inference time (ms) \\ % 列标题
    \midrule % 中间线条
    MotionFormer & 56 ($\pm$ 7)\\
    Post-Optimizer & 2616 ($\pm$ 292)\\
    Total & 2672 ($\pm$ 297)\\
    \bottomrule % 底部线条
  \end{tabular}
\end{table}

\subsection{Discussion}\label{subsec_discussion}
The application of RL in trajectory planning for urban automated driving is limited by the poor convergence of RL and the difficulty of designing reward functions. The proposed RITP method integrates RL with IL to overcome the convergence problem. It trains CriticFormer to evaluate the closed-loop effect of an action, according to which supervisory signals are generated for IL. In contrast to RL, the gradients computed by IL are sufficient to optimize deep neural networks such as MotionFormer. Then, MotionFormer trained through IL explores the state and action spaces using the trajectory noise, thus resolving the suboptimality and the online computational complexity issues in TRAVL. Moreover, Bayesian RewardFormer provides effective reward signals for RL. As indicated by the results presented in the previous subsections, RITP significantly outperforms the baselines in terms of the closed-loop metrics. Another important reason for the effectiveness of RITP is that CriticFormer, MotionFormer, and Bayesian RewardFormer utilize the attention mechanism to extract information from the query-centric representation effectively. We also found that combining data-driven and model-driven methods can further improve the performance of RITP. Due to the inevitable epistemic uncertainty in neural networks, RITP might plan dangerous trajectories in unfamiliar situations. The post-optimizer refines the trajectories outputted by RITP to correct these wrong plans.

It is worth noting that common IL-based methods can be incorporated into the proposed RITP framework by simply replacing MotionFormer module and its associated training objective. For example, MotionFormer can be replaced by UniAD model \cite{Hu_2023_CVPR}, which takes as input multi-view camera images, to enable end-to-end trajectory planning \cite{chen2023end}. By leveraging the proposed RL framework, the generalization ability of such incorporated models can be significantly enhanced for safe trajectory planning in complex urban scenarios. Beyond model replacement, the RITP framework's training objective could potentially be enhanced by integrating SOTA IL techniques. For instance, incorporating contrastive imitation learning, as employed in PLUTO \cite{cheng2024pluto}, holds potential for further performance improvement.

However, we only use 10000 scenarios, a small portion of the full dataset, in our experiments due to hardware limitations. The performance of RITP using more scenarios is worth investigating.

\section{Conclusion}\label{sec_conclusion}

This paper proposes a novel RL-based trajectory planning method for urban automated driving. The proposed method integrates IL into RL to improve the convergence, thus enabling planning for multiple future steps. Furthermore, a transformer-based Bayesian reward function is developed, which removes the linear structure assumption imposed by most existing methods and is applicable to RL in urban scenarios. Besides, a hybrid-driven trajectory planning framework is proposed to enhance the feasibility of the trajectories planned by the RL agent. The results of the experiments conducted on the large-scale real-world urban automated driving nuPlan dataset indicate that the proposed method significantly outperforms baselines with identical policy network structures and achieves competitive performance compared to the SOTA method in terms of closed-loop metrics.

\backmatter

\section*{Abbreviations}
\begin{tabularx}{\textwidth}{lX}
AD & Automated driving\\
AV(s) & Automated vehicle(s)\\
BC & Behavior cloning\\
BEV & Bird's-eye-view\\
IL & Imitation learning\\
IRL & Inverse reinforcement learning\\
LQR & Linear quadratic regulator\\
MDP & Markov decision process\\
MLP(s) & Multi-layer perceptron(s)\\
QCMAE & Query-centric masked autoencoder\\
RITP & Reinforced imitative trajectory\\& planning\\
RL & Reinforcement learning\\
TRAVL & Trajectory value learning\\

\end{tabularx}

\bmhead{Acknowledgements}

This work was supported by the National Natural Science Foundation of China (Grant No. 52472424) and the Fundamental Research Funds for the Central Universities (2023CDJXY-021).
%This work was supported by the ...

\section*{Declarations}

\textbf{Conflict of interest} On behalf of all authors, the corresponding author states that there is no conflict of interest.


\bibliographystyle{elsarticle-num-names}
\bibliography{sn-bibliography}% common bib file

\begin{thebibliography}{50}
\expandafter\ifx\csname natexlab\endcsname\relax\def\natexlab#1{#1}\fi
\providecommand{\url}[1]{\texttt{#1}}
\providecommand{\href}[2]{#2}
\providecommand{\path}[1]{#1}
\providecommand{\DOIprefix}{doi:}
\providecommand{\ArXivprefix}{arXiv:}
\providecommand{\URLprefix}{URL: }
\providecommand{\Pubmedprefix}{pmid:}
\providecommand{\doi}[1]{\href{http://dx.doi.org/#1}{\path{#1}}}
\providecommand{\Pubmed}[1]{\href{pmid:#1}{\path{#1}}}
\providecommand{\bibinfo}[2]{#2}
\ifx\xfnm\relax \def\xfnm[#1]{\unskip,\space#1}\fi
%Type = Article
\bibitem[{Bojarski et~al.(2016)Bojarski, Del~Testa, Dworakowski, Firner, Flepp,
  Goyal, Jackel, Monfort, Muller, Zhang et~al.}]{bojarski2016end}
\bibinfo{author}{M.~Bojarski}, \bibinfo{author}{D.~Del~Testa},
  \bibinfo{author}{D.~Dworakowski}, \bibinfo{author}{B.~Firner},
  \bibinfo{author}{B.~Flepp}, \bibinfo{author}{P.~Goyal},
  \bibinfo{author}{L.~D. Jackel}, \bibinfo{author}{M.~Monfort},
  \bibinfo{author}{U.~Muller}, \bibinfo{author}{J.~Zhang}, et~al.,
\newblock \bibinfo{title}{End to end learning for self-driving cars},
\newblock \bibinfo{journal}{arXiv preprint arXiv:1604.07316}
  (\bibinfo{year}{2016}).
%Type = Inproceedings
\bibitem[{Chen and Kr\"ahenb\"uhl(2022)}]{Chen_2022_CVPR}
\bibinfo{author}{D.~Chen}, \bibinfo{author}{P.~Kr\"ahenb\"uhl},
\newblock \bibinfo{title}{Learning from all vehicles},
\newblock in: \bibinfo{booktitle}{Proceedings of the IEEE/CVF Conference on
  Computer Vision and Pattern Recognition (CVPR)}, \bibinfo{year}{2022}, pp.
  \bibinfo{pages}{17222--17231}.
%Type = Inproceedings
\bibitem[{Hu et~al.(2023)Hu, Yang, Chen, Li, Sima, Zhu, Chai, Du, Lin, Wang,
  Lu, Jia, Liu, Dai, Qiao, and Li}]{Hu_2023_CVPR}
\bibinfo{author}{Y.~Hu}, \bibinfo{author}{J.~Yang}, \bibinfo{author}{L.~Chen},
  \bibinfo{author}{K.~Li}, \bibinfo{author}{C.~Sima}, \bibinfo{author}{X.~Zhu},
  \bibinfo{author}{S.~Chai}, \bibinfo{author}{S.~Du}, \bibinfo{author}{T.~Lin},
  \bibinfo{author}{W.~Wang}, \bibinfo{author}{L.~Lu}, \bibinfo{author}{X.~Jia},
  \bibinfo{author}{Q.~Liu}, \bibinfo{author}{J.~Dai},
  \bibinfo{author}{Y.~Qiao}, \bibinfo{author}{H.~Li},
\newblock \bibinfo{title}{Planning-oriented autonomous driving},
\newblock in: \bibinfo{booktitle}{Proceedings of the IEEE/CVF Conference on
  Computer Vision and Pattern Recognition (CVPR)}, \bibinfo{year}{2023}, pp.
  \bibinfo{pages}{17853--17862}.
%Type = Inproceedings
\bibitem[{Ross and Bagnell(2010)}]{pmlr-v9-ross10a}
\bibinfo{author}{S.~Ross}, \bibinfo{author}{D.~Bagnell},
\newblock \bibinfo{title}{Efficient reductions for imitation learning},
\newblock in: \bibinfo{editor}{Y.~W. Teh}, \bibinfo{editor}{M.~Titterington}
  (Eds.), \bibinfo{booktitle}{Proceedings of the Thirteenth International
  Conference on Artificial Intelligence and Statistics},
  volume~\bibinfo{volume}{9} of \textit{\bibinfo{series}{Proceedings of Machine
  Learning Research}}, \bibinfo{publisher}{PMLR}, \bibinfo{address}{Chia Laguna
  Resort, Sardinia, Italy}, \bibinfo{year}{2010}, pp.
  \bibinfo{pages}{661--668}.
%Type = Inproceedings
\bibitem[{Ross et~al.(2011)Ross, Gordon, and Bagnell}]{pmlr-v15-ross11a}
\bibinfo{author}{S.~Ross}, \bibinfo{author}{G.~Gordon},
  \bibinfo{author}{D.~Bagnell},
\newblock \bibinfo{title}{A reduction of imitation learning and structured
  prediction to no-regret online learning},
\newblock in: \bibinfo{editor}{G.~Gordon}, \bibinfo{editor}{D.~Dunson},
  \bibinfo{editor}{M.~Dud\'ik} (Eds.), \bibinfo{booktitle}{Proceedings of the
  Fourteenth International Conference on Artificial Intelligence and
  Statistics}, volume~\bibinfo{volume}{15} of
  \textit{\bibinfo{series}{Proceedings of Machine Learning Research}},
  \bibinfo{publisher}{PMLR}, \bibinfo{address}{Fort Lauderdale, FL, USA},
  \bibinfo{year}{2011}, pp. \bibinfo{pages}{627--635}.
%Type = Article
\bibitem[{Zhang and Cho(2017)}]{Zhang_Cho_2017}
\bibinfo{author}{J.~Zhang}, \bibinfo{author}{K.~Cho},
\newblock \bibinfo{title}{Query-efficient imitation learning for end-to-end
  simulated driving},
\newblock \bibinfo{journal}{Proceedings of the AAAI Conference on Artificial
  Intelligence} \bibinfo{volume}{31} (\bibinfo{year}{2017}).
  \DOIprefix\doi{10.1609/aaai.v31i1.10857}.
%Type = Inproceedings
\bibitem[{Zhang et~al.(2022)Zhang, Guo, Zeng, Xiong, Dai, Hu, Ren, and
  Urtasun}]{Zhang_Guo_2022}
\bibinfo{author}{C.~Zhang}, \bibinfo{author}{R.~Guo},
  \bibinfo{author}{W.~Zeng}, \bibinfo{author}{Y.~Xiong},
  \bibinfo{author}{B.~Dai}, \bibinfo{author}{R.~Hu}, \bibinfo{author}{M.~Ren},
  \bibinfo{author}{R.~Urtasun},
\newblock \bibinfo{title}{Rethinking closed-loop training for autonomous
  driving},
\newblock in: \bibinfo{editor}{S.~Avidan}, \bibinfo{editor}{G.~Brostow},
  \bibinfo{editor}{M.~Ciss{\'e}}, \bibinfo{editor}{G.~M. Farinella},
  \bibinfo{editor}{T.~Hassner} (Eds.), \bibinfo{booktitle}{Computer Vision --
  ECCV 2022}, \bibinfo{publisher}{Springer Nature Switzerland},
  \bibinfo{address}{Cham}, \bibinfo{year}{2022}, pp. \bibinfo{pages}{264--282}.
%Type = Article
\bibitem[{Sallab et~al.(2016)Sallab, Abdou, Perot, and
  Yogamani}]{sallab2016end}
\bibinfo{author}{A.~E. Sallab}, \bibinfo{author}{M.~Abdou},
  \bibinfo{author}{E.~Perot}, \bibinfo{author}{S.~Yogamani},
\newblock \bibinfo{title}{End-to-end deep reinforcement learning for lane
  keeping assist},
\newblock \bibinfo{journal}{arXiv preprint arXiv:1612.04340}
  (\bibinfo{year}{2016}).
%Type = Inproceedings
\bibitem[{Kendall et~al.(2019)Kendall, Hawke, Janz, Mazur, Reda, Allen, Lam,
  Bewley, and Shah}]{8793742}
\bibinfo{author}{A.~Kendall}, \bibinfo{author}{J.~Hawke},
  \bibinfo{author}{D.~Janz}, \bibinfo{author}{P.~Mazur},
  \bibinfo{author}{D.~Reda}, \bibinfo{author}{J.-M. Allen},
  \bibinfo{author}{V.-D. Lam}, \bibinfo{author}{A.~Bewley},
  \bibinfo{author}{A.~Shah},
\newblock \bibinfo{title}{Learning to drive in a day},
\newblock in: \bibinfo{booktitle}{2019 International Conference on Robotics and
  Automation (ICRA)}, \bibinfo{year}{2019}, pp. \bibinfo{pages}{8248--8254}.
  \DOIprefix\doi{10.1109/ICRA.2019.8793742}.
%Type = Article
\bibitem[{Li et~al.(2023)Li, Qiu, Yang, Li, Li, Chu, Green, and Li}]{9978654}
\bibinfo{author}{G.~Li}, \bibinfo{author}{Y.~Qiu}, \bibinfo{author}{Y.~Yang},
  \bibinfo{author}{Z.~Li}, \bibinfo{author}{S.~Li}, \bibinfo{author}{W.~Chu},
  \bibinfo{author}{P.~Green}, \bibinfo{author}{S.~E. Li},
\newblock \bibinfo{title}{Lane change strategies for autonomous vehicles: A
  deep reinforcement learning approach based on transformer},
\newblock \bibinfo{journal}{IEEE Transactions on Intelligent Vehicles}
  \bibinfo{volume}{8} (\bibinfo{year}{2023}) \bibinfo{pages}{2197--2211}.
  \DOIprefix\doi{10.1109/TIV.2022.3227921}.
%Type = Article
\bibitem[{Wang et~al.(2022)Wang, Hu, Li, and Li}]{9325948}
\bibinfo{author}{G.~Wang}, \bibinfo{author}{J.~Hu}, \bibinfo{author}{Z.~Li},
  \bibinfo{author}{L.~Li},
\newblock \bibinfo{title}{Harmonious lane changing via deep reinforcement
  learning},
\newblock \bibinfo{journal}{IEEE Transactions on Intelligent Transportation
  Systems} \bibinfo{volume}{23} (\bibinfo{year}{2022})
  \bibinfo{pages}{4642--4650}. \DOIprefix\doi{10.1109/TITS.2020.3047129}.
%Type = Inproceedings
\bibitem[{Wang and Chan(2017)}]{8317735}
\bibinfo{author}{P.~Wang}, \bibinfo{author}{C.-Y. Chan},
\newblock \bibinfo{title}{Formulation of deep reinforcement learning
  architecture toward autonomous driving for on-ramp merge},
\newblock in: \bibinfo{booktitle}{2017 IEEE 20th International Conference on
  Intelligent Transportation Systems (ITSC)}, \bibinfo{year}{2017}, pp.
  \bibinfo{pages}{1--6}. \DOIprefix\doi{10.1109/ITSC.2017.8317735}.
%Type = Article
\bibitem[{Chen et~al.(2024)Chen, Wu, Chitta, Jaeger, Geiger, and
  Li}]{chen2023end}
\bibinfo{author}{L.~Chen}, \bibinfo{author}{P.~Wu},
  \bibinfo{author}{K.~Chitta}, \bibinfo{author}{B.~Jaeger},
  \bibinfo{author}{A.~Geiger}, \bibinfo{author}{H.~Li},
\newblock \bibinfo{title}{End-to-end autonomous driving: Challenges and
  frontiers},
\newblock \bibinfo{journal}{IEEE Transactions on Pattern Analysis and Machine
  Intelligence} \bibinfo{volume}{46} (\bibinfo{year}{2024})
  \bibinfo{pages}{10164--10183}. \DOIprefix\doi{10.1109/TPAMI.2024.3435937}.
%Type = Article
\bibitem[{Huang et~al.(2023)Huang, Wu, and Lv}]{9694460}
\bibinfo{author}{Z.~Huang}, \bibinfo{author}{J.~Wu}, \bibinfo{author}{C.~Lv},
\newblock \bibinfo{title}{Efficient deep reinforcement learning with imitative
  expert priors for autonomous driving},
\newblock \bibinfo{journal}{IEEE Transactions on Neural Networks and Learning
  Systems} \bibinfo{volume}{34} (\bibinfo{year}{2023})
  \bibinfo{pages}{7391--7403}. \DOIprefix\doi{10.1109/TNNLS.2022.3142822}.
%Type = Inproceedings
\bibitem[{Zhang et~al.(2021)Zhang, Liniger, Dai, Yu, and
  Van~Gool}]{Zhang_2021_ICCV}
\bibinfo{author}{Z.~Zhang}, \bibinfo{author}{A.~Liniger},
  \bibinfo{author}{D.~Dai}, \bibinfo{author}{F.~Yu},
  \bibinfo{author}{L.~Van~Gool},
\newblock \bibinfo{title}{End-to-end urban driving by imitating a reinforcement
  learning coach},
\newblock in: \bibinfo{booktitle}{Proceedings of the IEEE/CVF International
  Conference on Computer Vision (ICCV)}, \bibinfo{year}{2021}, pp.
  \bibinfo{pages}{15222--15232}.
%Type = Article
\bibitem[{Cusumano-Towner et~al.(2025)Cusumano-Towner, Hafner, Hertzberg,
  Huval, Petrenko, Vinitsky, Wijmans, Killian, Bowers, Sener
  et~al.}]{cusumano2025robust}
\bibinfo{author}{M.~Cusumano-Towner}, \bibinfo{author}{D.~Hafner},
  \bibinfo{author}{A.~Hertzberg}, \bibinfo{author}{B.~Huval},
  \bibinfo{author}{A.~Petrenko}, \bibinfo{author}{E.~Vinitsky},
  \bibinfo{author}{E.~Wijmans}, \bibinfo{author}{T.~Killian},
  \bibinfo{author}{S.~Bowers}, \bibinfo{author}{O.~Sener}, et~al.,
\newblock \bibinfo{title}{Robust autonomy emerges from self-play},
\newblock \bibinfo{journal}{arXiv preprint arXiv:2502.03349}
  (\bibinfo{year}{2025}).
%Type = Article
\bibitem[{Zhu and Zhao(2022)}]{9660769}
\bibinfo{author}{Z.~Zhu}, \bibinfo{author}{H.~Zhao},
\newblock \bibinfo{title}{A survey of deep rl and il for autonomous driving
  policy learning},
\newblock \bibinfo{journal}{IEEE Transactions on Intelligent Transportation
  Systems} \bibinfo{volume}{23} (\bibinfo{year}{2022})
  \bibinfo{pages}{14043--14065}. \DOIprefix\doi{10.1109/TITS.2021.3134702}.
%Type = Article
\bibitem[{Kiran et~al.(2022)Kiran, Sobh, Talpaert, Mannion, Sallab, Yogamani,
  and P\'erez}]{9351818}
\bibinfo{author}{B.~R. Kiran}, \bibinfo{author}{I.~Sobh},
  \bibinfo{author}{V.~Talpaert}, \bibinfo{author}{P.~Mannion},
  \bibinfo{author}{A.~A.~A. Sallab}, \bibinfo{author}{S.~Yogamani},
  \bibinfo{author}{P.~P\'erez},
\newblock \bibinfo{title}{Deep reinforcement learning for autonomous driving: A
  survey},
\newblock \bibinfo{journal}{IEEE Transactions on Intelligent Transportation
  Systems} \bibinfo{volume}{23} (\bibinfo{year}{2022})
  \bibinfo{pages}{4909--4926}. \DOIprefix\doi{10.1109/TITS.2021.3054625}.
%Type = Article
\bibitem[{Li et~al.(2025)Li, Ren, Wang, Wen, Li, Xu, Zhan, Xia, Jia, Lang
  et~al.}]{li2025finetuning}
\bibinfo{author}{D.~Li}, \bibinfo{author}{J.~Ren}, \bibinfo{author}{Y.~Wang},
  \bibinfo{author}{X.~Wen}, \bibinfo{author}{P.~Li}, \bibinfo{author}{L.~Xu},
  \bibinfo{author}{K.~Zhan}, \bibinfo{author}{Z.~Xia},
  \bibinfo{author}{P.~Jia}, \bibinfo{author}{X.~Lang}, et~al.,
\newblock \bibinfo{title}{Finetuning generative trajectory model with
  reinforcement learning from human feedback},
\newblock \bibinfo{journal}{arXiv preprint arXiv:2503.10434}
  (\bibinfo{year}{2025}).
%Type = Article
\bibitem[{Sharifzadeh et~al.(2016)Sharifzadeh, Chiotellis, Triebel, and
  Cremers}]{sharifzadeh2016learning}
\bibinfo{author}{S.~Sharifzadeh}, \bibinfo{author}{I.~Chiotellis},
  \bibinfo{author}{R.~Triebel}, \bibinfo{author}{D.~Cremers},
\newblock \bibinfo{title}{Learning to drive using inverse reinforcement
  learning and deep q-networks},
\newblock \bibinfo{journal}{arXiv preprint arXiv:1612.03653}
  (\bibinfo{year}{2016}).
%Type = Article
\bibitem[{Wu et~al.(2020)Wu, Sun, Zhan, Yang, and Tomizuka}]{9126156}
\bibinfo{author}{Z.~Wu}, \bibinfo{author}{L.~Sun}, \bibinfo{author}{W.~Zhan},
  \bibinfo{author}{C.~Yang}, \bibinfo{author}{M.~Tomizuka},
\newblock \bibinfo{title}{Efficient sampling-based maximum entropy inverse
  reinforcement learning with application to autonomous driving},
\newblock \bibinfo{journal}{IEEE Robotics and Automation Letters}
  \bibinfo{volume}{5} (\bibinfo{year}{2020}) \bibinfo{pages}{5355--5362}.
  \DOIprefix\doi{10.1109/LRA.2020.3005126}.
%Type = Article
\bibitem[{Huang et~al.(2022)Huang, Wu, and Lv}]{9460807}
\bibinfo{author}{Z.~Huang}, \bibinfo{author}{J.~Wu}, \bibinfo{author}{C.~Lv},
\newblock \bibinfo{title}{Driving behavior modeling using naturalistic human
  driving data with inverse reinforcement learning},
\newblock \bibinfo{journal}{IEEE Transactions on Intelligent Transportation
  Systems} \bibinfo{volume}{23} (\bibinfo{year}{2022})
  \bibinfo{pages}{10239--10251}. \DOIprefix\doi{10.1109/TITS.2021.3088935}.
%Type = Article
\bibitem[{Caesar et~al.(2021)Caesar, Kabzan, Tan, Fong, Wolff, Lang, Fletcher,
  Beijbom, and Omari}]{caesar2021nuplan}
\bibinfo{author}{H.~Caesar}, \bibinfo{author}{J.~Kabzan},
  \bibinfo{author}{K.~S. Tan}, \bibinfo{author}{W.~K. Fong},
  \bibinfo{author}{E.~Wolff}, \bibinfo{author}{A.~Lang},
  \bibinfo{author}{L.~Fletcher}, \bibinfo{author}{O.~Beijbom},
  \bibinfo{author}{S.~Omari},
\newblock \bibinfo{title}{nuplan: A closed-loop ml-based planning benchmark for
  autonomous vehicles},
\newblock \bibinfo{journal}{arXiv preprint arXiv:2106.11810}
  (\bibinfo{year}{2021}).
%Type = Article
\bibitem[{Karnchanachari et~al.(2024)Karnchanachari, Geromichalos, Tan, Li,
  Eriksen, Yaghoubi, Mehdipour, Bernasconi, Fong, Guo
  et~al.}]{karnchanachari2024towards}
\bibinfo{author}{N.~Karnchanachari}, \bibinfo{author}{D.~Geromichalos},
  \bibinfo{author}{K.~S. Tan}, \bibinfo{author}{N.~Li},
  \bibinfo{author}{C.~Eriksen}, \bibinfo{author}{S.~Yaghoubi},
  \bibinfo{author}{N.~Mehdipour}, \bibinfo{author}{G.~Bernasconi},
  \bibinfo{author}{W.~K. Fong}, \bibinfo{author}{Y.~Guo}, et~al.,
\newblock \bibinfo{title}{Towards learning-based planning: The nuplan benchmark
  for real-world autonomous driving},
\newblock \bibinfo{journal}{arXiv preprint arXiv:2403.04133}
  (\bibinfo{year}{2024}).
%Type = Article
\bibitem[{Sutton(1988)}]{sutton1988learning}
\bibinfo{author}{R.~S. Sutton},
\newblock \bibinfo{title}{Learning to predict by the methods of temporal
  differences},
\newblock \bibinfo{journal}{Machine learning} \bibinfo{volume}{3}
  (\bibinfo{year}{1988}) \bibinfo{pages}{9--44}.
%Type = Article
\bibitem[{Mnih et~al.(2015)Mnih, Kavukcuoglu, Silver, Rusu, Veness, Bellemare,
  Graves, Riedmiller, Fidjeland, Ostrovski et~al.}]{mnih2015human}
\bibinfo{author}{V.~Mnih}, \bibinfo{author}{K.~Kavukcuoglu},
  \bibinfo{author}{D.~Silver}, \bibinfo{author}{A.~A. Rusu},
  \bibinfo{author}{J.~Veness}, \bibinfo{author}{M.~G. Bellemare},
  \bibinfo{author}{A.~Graves}, \bibinfo{author}{M.~Riedmiller},
  \bibinfo{author}{A.~K. Fidjeland}, \bibinfo{author}{G.~Ostrovski}, et~al.,
\newblock \bibinfo{title}{Human-level control through deep reinforcement
  learning},
\newblock \bibinfo{journal}{nature} \bibinfo{volume}{518}
  (\bibinfo{year}{2015}) \bibinfo{pages}{529--533}.
  \DOIprefix\doi{10.1038/nature14236}.
%Type = Inproceedings
\bibitem[{Fujimoto et~al.(2018)Fujimoto, van Hoof, and
  Meger}]{pmlr-v80-fujimoto18a}
\bibinfo{author}{S.~Fujimoto}, \bibinfo{author}{H.~van Hoof},
  \bibinfo{author}{D.~Meger},
\newblock \bibinfo{title}{Addressing function approximation error in
  actor-critic methods},
\newblock in: \bibinfo{editor}{J.~Dy}, \bibinfo{editor}{A.~Krause} (Eds.),
  \bibinfo{booktitle}{Proceedings of the 35th International Conference on
  Machine Learning}, volume~\bibinfo{volume}{80} of
  \textit{\bibinfo{series}{Proceedings of Machine Learning Research}},
  \bibinfo{publisher}{PMLR}, \bibinfo{year}{2018}, pp.
  \bibinfo{pages}{1587--1596}.
%Type = Inproceedings
\bibitem[{Liu et~al.(2021)Liu, Yang, Huang, Li, Dang, and Li}]{9517448}
\bibinfo{author}{J.~Liu}, \bibinfo{author}{Z.~Yang},
  \bibinfo{author}{Z.~Huang}, \bibinfo{author}{W.~Li},
  \bibinfo{author}{S.~Dang}, \bibinfo{author}{H.~Li},
\newblock \bibinfo{title}{Simulation performance evaluation of pure pursuit,
  stanley, lqr, mpc controller for autonomous vehicles},
\newblock in: \bibinfo{booktitle}{2021 IEEE International Conference on
  Real-time Computing and Robotics (RCAR)}, \bibinfo{year}{2021}, pp.
  \bibinfo{pages}{1444--1449}.
%Type = Inproceedings
\bibitem[{Zhou et~al.(2023)Zhou, Wang, Li, and Huang}]{Zhou_2023_CVPR}
\bibinfo{author}{Z.~Zhou}, \bibinfo{author}{J.~Wang}, \bibinfo{author}{Y.-H.
  Li}, \bibinfo{author}{Y.-K. Huang},
\newblock \bibinfo{title}{Query-centric trajectory prediction},
\newblock in: \bibinfo{booktitle}{Proceedings of the IEEE/CVF Conference on
  Computer Vision and Pattern Recognition (CVPR)}, \bibinfo{year}{2023}, pp.
  \bibinfo{pages}{17863--17873}.
%Type = Inproceedings
\bibitem[{Tancik et~al.(2020)Tancik, Srinivasan, Mildenhall, Fridovich-Keil,
  Raghavan, Singhal, Ramamoorthi, Barron, and Ng}]{NEURIPS2020_55053683}
\bibinfo{author}{M.~Tancik}, \bibinfo{author}{P.~Srinivasan},
  \bibinfo{author}{B.~Mildenhall}, \bibinfo{author}{S.~Fridovich-Keil},
  \bibinfo{author}{N.~Raghavan}, \bibinfo{author}{U.~Singhal},
  \bibinfo{author}{R.~Ramamoorthi}, \bibinfo{author}{J.~Barron},
  \bibinfo{author}{R.~Ng},
\newblock \bibinfo{title}{Fourier features let networks learn high frequency
  functions in low dimensional domains},
\newblock in: \bibinfo{editor}{H.~Larochelle}, \bibinfo{editor}{M.~Ranzato},
  \bibinfo{editor}{R.~Hadsell}, \bibinfo{editor}{M.~Balcan},
  \bibinfo{editor}{H.~Lin} (Eds.), \bibinfo{booktitle}{Advances in Neural
  Information Processing Systems}, volume~\bibinfo{volume}{33},
  \bibinfo{publisher}{Curran Associates, Inc.}, \bibinfo{year}{2020}, pp.
  \bibinfo{pages}{7537--7547}.
%Type = Inproceedings
\bibitem[{Werling et~al.(2010)Werling, Ziegler, Kammel, and Thrun}]{5509799}
\bibinfo{author}{M.~Werling}, \bibinfo{author}{J.~Ziegler},
  \bibinfo{author}{S.~Kammel}, \bibinfo{author}{S.~Thrun},
\newblock \bibinfo{title}{Optimal trajectory generation for dynamic street
  scenarios in a fren\'et frame},
\newblock in: \bibinfo{booktitle}{2010 IEEE International Conference on
  Robotics and Automation}, \bibinfo{year}{2010}, pp.
  \bibinfo{pages}{987--993}. \DOIprefix\doi{10.1109/ROBOT.2010.5509799}.
%Type = Inproceedings
\bibitem[{Cho et~al.(2014)Cho, van Merri{\"e}nboer, Gulcehre, Bahdanau,
  Bougares, Schwenk, and Bengio}]{cho-etal-2014-learning}
\bibinfo{author}{K.~Cho}, \bibinfo{author}{B.~van Merri{\"e}nboer},
  \bibinfo{author}{C.~Gulcehre}, \bibinfo{author}{D.~Bahdanau},
  \bibinfo{author}{F.~Bougares}, \bibinfo{author}{H.~Schwenk},
  \bibinfo{author}{Y.~Bengio},
\newblock \bibinfo{title}{Learning phrase representations using {RNN}
  encoder{--}decoder for statistical machine translation},
\newblock in: \bibinfo{editor}{A.~Moschitti}, \bibinfo{editor}{B.~Pang},
  \bibinfo{editor}{W.~Daelemans} (Eds.), \bibinfo{booktitle}{Proceedings of the
  2014 Conference on Empirical Methods in Natural Language Processing
  ({EMNLP})}, \bibinfo{publisher}{Association for Computational Linguistics},
  \bibinfo{address}{Doha, Qatar}, \bibinfo{year}{2014}, pp.
  \bibinfo{pages}{1724--1734}. \DOIprefix\doi{10.3115/v1/D14-1179}.
%Type = Article
\bibitem[{Zeng et~al.(2024)Zeng, Zheng, Yang, and Li}]{AVRL}
\bibinfo{author}{D.~Zeng}, \bibinfo{author}{L.~Zheng},
  \bibinfo{author}{X.~Yang}, \bibinfo{author}{Y.~Li},
\newblock \bibinfo{title}{Uncertainty-aware human-like driving policy learning
  with deep bayesian inverse reinforcement learning},
\newblock \bibinfo{journal}{Transportmetrica A: Transport Science}
  \bibinfo{volume}{0} (\bibinfo{year}{2024}) \bibinfo{pages}{1--25}.
  \DOIprefix\doi{10.1080/23249935.2024.2318621}.
%Type = Book
\bibitem[{Neal(2012)}]{neal2012bayesian}
\bibinfo{author}{R.~M. Neal}, \bibinfo{title}{Bayesian Learning for Neural
  Networks}, volume \bibinfo{volume}{118}, \bibinfo{publisher}{Springer},
  \bibinfo{address}{New York, NY}, \bibinfo{year}{2012}.
%Type = Article
\bibitem[{Blei et~al.(2017)Blei, Kucukelbir, and
  McAuliffe}]{blei2017variational}
\bibinfo{author}{D.~M. Blei}, \bibinfo{author}{A.~Kucukelbir},
  \bibinfo{author}{J.~D. McAuliffe},
\newblock \bibinfo{title}{Variational inference: A review for statisticians},
\newblock \bibinfo{journal}{Journal of the American Statistical Association}
  \bibinfo{volume}{112} (\bibinfo{year}{2017}) \bibinfo{pages}{859--877}.
%Type = Article
\bibitem[{Hinton et~al.(2012)Hinton, Srivastava, Krizhevsky, Sutskever, and
  Salakhutdinov}]{hinton2012improving}
\bibinfo{author}{G.~E. Hinton}, \bibinfo{author}{N.~Srivastava},
  \bibinfo{author}{A.~Krizhevsky}, \bibinfo{author}{I.~Sutskever},
  \bibinfo{author}{R.~R. Salakhutdinov},
\newblock \bibinfo{title}{Improving neural networks by preventing co-adaptation
  of feature detectors},
\newblock \bibinfo{journal}{arXiv preprint arXiv:1207.0580}
  (\bibinfo{year}{2012}).
%Type = Inproceedings
\bibitem[{Gal and Ghahramani(2016)}]{gal2016dropout}
\bibinfo{author}{Y.~Gal}, \bibinfo{author}{Z.~Ghahramani},
\newblock \bibinfo{title}{Dropout as a bayesian approximation: Representing
  model uncertainty in deep learning},
\newblock in: \bibinfo{booktitle}{International Conference on Machine
  Learning}, \bibinfo{organization}{PMLR}, \bibinfo{year}{2016}, pp.
  \bibinfo{pages}{1050--1059}.
%Type = Article
\bibitem[{Gal(2016)}]{gal2016uncertainty}
\bibinfo{author}{Y.~Gal},
\newblock \bibinfo{title}{Uncertainty in deep learning}
  (\bibinfo{year}{2016}).
%Type = Inproceedings
\bibitem[{Dauner et~al.(2023)Dauner, Hallgarten, Geiger, and
  Chitta}]{pmlr-v229-dauner23a}
\bibinfo{author}{D.~Dauner}, \bibinfo{author}{M.~Hallgarten},
  \bibinfo{author}{A.~Geiger}, \bibinfo{author}{K.~Chitta},
\newblock \bibinfo{title}{Parting with misconceptions about learning-based
  vehicle motion planning},
\newblock in: \bibinfo{editor}{J.~Tan}, \bibinfo{editor}{M.~Toussaint},
  \bibinfo{editor}{K.~Darvish} (Eds.), \bibinfo{booktitle}{Proceedings of The
  7th Conference on Robot Learning}, volume \bibinfo{volume}{229} of
  \textit{\bibinfo{series}{Proceedings of Machine Learning Research}},
  \bibinfo{publisher}{PMLR}, \bibinfo{year}{2023}, pp.
  \bibinfo{pages}{1268--1281}.
%Type = Article
\bibitem[{Treiber et~al.(2000)Treiber, Hennecke, and
  Helbing}]{PhysRevE.62.1805}
\bibinfo{author}{M.~Treiber}, \bibinfo{author}{A.~Hennecke},
  \bibinfo{author}{D.~Helbing},
\newblock \bibinfo{title}{Congested traffic states in empirical observations
  and microscopic simulations},
\newblock \bibinfo{journal}{Phys. Rev. E} \bibinfo{volume}{62}
  (\bibinfo{year}{2000}) \bibinfo{pages}{1805--1824}.
  \DOIprefix\doi{10.1103/PhysRevE.62.1805}.
%Type = Inproceedings
\bibitem[{Scheel et~al.(2022)Scheel, Bergamini, Wolczyk, Osi\'nski, and
  Ondruska}]{UrbanDriver}
\bibinfo{author}{O.~Scheel}, \bibinfo{author}{L.~Bergamini},
  \bibinfo{author}{M.~Wolczyk}, \bibinfo{author}{B.~Osi\'nski},
  \bibinfo{author}{P.~Ondruska},
\newblock \bibinfo{title}{Urban driver: Learning to drive from real-world
  demonstrations using policy gradients},
\newblock in: \bibinfo{editor}{A.~Faust}, \bibinfo{editor}{D.~Hsu},
  \bibinfo{editor}{G.~Neumann} (Eds.), \bibinfo{booktitle}{Proceedings of the
  5th Conference on Robot Learning}, volume \bibinfo{volume}{164} of
  \textit{\bibinfo{series}{Proceedings of Machine Learning Research}},
  \bibinfo{publisher}{PMLR}, \bibinfo{year}{2022}, pp.
  \bibinfo{pages}{718--728}.
%Type = Article
\bibitem[{Cheng et~al.(2024)Cheng, Chen, and Chen}]{cheng2024pluto}
\bibinfo{author}{J.~Cheng}, \bibinfo{author}{Y.~Chen},
  \bibinfo{author}{Q.~Chen},
\newblock \bibinfo{title}{Pluto: Pushing the limit of imitation learning-based
  planning for autonomous driving},
\newblock \bibinfo{journal}{arXiv preprint arXiv:2404.14327}
  (\bibinfo{year}{2024}).
%Type = Inproceedings
\bibitem[{Carion et~al.(2020)Carion, Massa, Synnaeve, Usunier, Kirillov, and
  Zagoruyko}]{DETR}
\bibinfo{author}{N.~Carion}, \bibinfo{author}{F.~Massa},
  \bibinfo{author}{G.~Synnaeve}, \bibinfo{author}{N.~Usunier},
  \bibinfo{author}{A.~Kirillov}, \bibinfo{author}{S.~Zagoruyko},
\newblock \bibinfo{title}{End-to-end object detection with transformers},
\newblock in: \bibinfo{editor}{A.~Vedaldi}, \bibinfo{editor}{H.~Bischof},
  \bibinfo{editor}{T.~Brox}, \bibinfo{editor}{J.-M. Frahm} (Eds.),
  \bibinfo{booktitle}{Computer Vision -- ECCV 2020},
  \bibinfo{publisher}{Springer International Publishing},
  \bibinfo{address}{Cham}, \bibinfo{year}{2020}, pp. \bibinfo{pages}{213--229}.
%Type = Inproceedings
\bibitem[{Lan et~al.(2024)Lan, Jiang, Mu, Chen, and Li}]{lan2024sept}
\bibinfo{author}{Z.~Lan}, \bibinfo{author}{Y.~Jiang}, \bibinfo{author}{Y.~Mu},
  \bibinfo{author}{C.~Chen}, \bibinfo{author}{S.~E. Li},
\newblock \bibinfo{title}{{SEPT}: Towards efficient scene representation
  learning for motion prediction},
\newblock in: \bibinfo{booktitle}{The Twelfth International Conference on
  Learning Representations}, \bibinfo{year}{2024}.
%Type = Inproceedings
\bibitem[{Chen et~al.(2017)Chen, Choi, Yu, Han, and
  Chandraker}]{NIPS2017_e1e32e23}
\bibinfo{author}{G.~Chen}, \bibinfo{author}{W.~Choi}, \bibinfo{author}{X.~Yu},
  \bibinfo{author}{T.~Han}, \bibinfo{author}{M.~Chandraker},
\newblock \bibinfo{title}{Learning efficient object detection models with
  knowledge distillation},
\newblock in: \bibinfo{editor}{I.~Guyon}, \bibinfo{editor}{U.~V. Luxburg},
  \bibinfo{editor}{S.~Bengio}, \bibinfo{editor}{H.~Wallach},
  \bibinfo{editor}{R.~Fergus}, \bibinfo{editor}{S.~Vishwanathan},
  \bibinfo{editor}{R.~Garnett} (Eds.), \bibinfo{booktitle}{Advances in Neural
  Information Processing Systems}, volume~\bibinfo{volume}{30},
  \bibinfo{publisher}{Curran Associates, Inc.}, \bibinfo{year}{2017}.
%Type = Inproceedings
\bibitem[{Shen et~al.(2022)Shen, Xu, Yang, Li, and Guo}]{Shen_2022_CVPR}
\bibinfo{author}{Y.~Shen}, \bibinfo{author}{L.~Xu}, \bibinfo{author}{Y.~Yang},
  \bibinfo{author}{Y.~Li}, \bibinfo{author}{Y.~Guo},
\newblock \bibinfo{title}{Self-distillation from the last mini-batch for
  consistency regularization},
\newblock in: \bibinfo{booktitle}{Proceedings of the IEEE/CVF Conference on
  Computer Vision and Pattern Recognition (CVPR)}, \bibinfo{year}{2022}, pp.
  \bibinfo{pages}{11943--11952}.
%Type = Article
\bibitem[{Kingma and Ba(2014)}]{adam}
\bibinfo{author}{D.~P. Kingma}, \bibinfo{author}{J.~Ba},
\newblock \bibinfo{title}{Adam: A method for stochastic optimization},
\newblock \bibinfo{journal}{arXiv preprint arXiv:1412.6980}
  (\bibinfo{year}{2014}).
%Type = Article
\bibitem[{Loshchilov and Hutter(2017)}]{adamw}
\bibinfo{author}{I.~Loshchilov}, \bibinfo{author}{F.~Hutter},
\newblock \bibinfo{title}{Decoupled weight decay regularization},
\newblock \bibinfo{journal}{arXiv preprint arXiv:1711.05101}
  (\bibinfo{year}{2017}).
%Type = Inproceedings
\bibitem[{Smith and Topin(2019)}]{onecyclelr}
\bibinfo{author}{L.~N. Smith}, \bibinfo{author}{N.~Topin},
\newblock \bibinfo{title}{{Super-convergence: very fast training of neural
  networks using large learning rates}},
\newblock in: \bibinfo{editor}{T.~Pham} (Ed.), \bibinfo{booktitle}{Artificial
  Intelligence and Machine Learning for Multi-Domain Operations Applications},
  volume \bibinfo{volume}{11006}, \bibinfo{organization}{International Society
  for Optics and Photonics}, \bibinfo{publisher}{SPIE}, \bibinfo{year}{2019},
  p. \bibinfo{pages}{1100612}. \DOIprefix\doi{10.1117/12.2520589}.
%Type = Article
\bibitem[{Fan et~al.(2018)Fan, Zhu, Liu, Zhang, Zhuang, Li, Zhu, Hu, Li, and
  Kong}]{fan2018baidu}
\bibinfo{author}{H.~Fan}, \bibinfo{author}{F.~Zhu}, \bibinfo{author}{C.~Liu},
  \bibinfo{author}{L.~Zhang}, \bibinfo{author}{L.~Zhuang},
  \bibinfo{author}{D.~Li}, \bibinfo{author}{W.~Zhu}, \bibinfo{author}{J.~Hu},
  \bibinfo{author}{H.~Li}, \bibinfo{author}{Q.~Kong},
\newblock \bibinfo{title}{Baidu apollo em motion planner},
\newblock \bibinfo{journal}{arXiv preprint arXiv:1807.08048}
  (\bibinfo{year}{2018}).

\end{thebibliography}

\end{document}